\documentclass[UTF8]{IEEEtran}
\usepackage{cite}
\usepackage{amsmath,amssymb,amsfonts}
\usepackage{algorithmic}
\usepackage{graphicx}
\usepackage{textcomp}
\usepackage{float}
\usepackage{algorithm}
\usepackage{algorithmic}
\usepackage{multirow}
\usepackage{booktabs} 
\usepackage{tikz}
\usepackage{calc}
\usepackage{bbding}
\usepackage{pifont}
\usepackage{wasysym}
\usepackage{amssymb}
\usepackage{color}
\usepackage{subfigure}
\usepackage{graphicx}
\usepackage{geometry}
\usepackage{pgfplots}
\usepackage{tikz}
\usepackage[switch]{lineno}

\begin{document}
\title{Three-Stage Cascade Framework for Blurry Video Frame Interpolation}
\author{Pengcheng Lei, Zaoming Yan, Tingting Wang, Faming Fang and Guixu Zhang
\thanks{Pengcheng Lei, Zaoming Yan, Tingting Wang, Faming Fang and Guixu Zhang are with  the School of Computer Science and Technology, East China Normal University, Shanghai 200062, China (e-mail: pengchenglei1995@163.com; fmfang@cs.ecnu.edu.cn).
}
}

\maketitle

\begin{abstract}
	Blurry video frame interpolation (BVFI) aims to generate high-frame-rate clear videos from low-frame-rate blurry videos, is a challenging but important topic in the computer vision community. Blurry videos not only provide spatial and temporal information like clear videos, but also contain additional motion information hidden in each blurry frame. However, existing BVFI methods usually fail to fully leverage all valuable information, which ultimately hinders their performance. In this paper, we propose a simple end-to-end three-stage framework to fully explore useful information from blurry videos. The frame interpolation stage designs a temporal deformable network to directly sample useful information from blurry inputs and synthesize an intermediate frame at an arbitrary time interval. The temporal feature fusion stage explores the long-term temporal information for each target frame through a bi-directional recurrent deformable alignment network. And the deblurring stage applies a transformer-empowered Taylor approximation network to recursively recover the high-frequency details. The proposed three-stage framework has clear task assignment for each module and offers good expandability, the effectiveness of which are demonstrated by various experimental results. We evaluate our model on four benchmarks, including the Adobe240 dataset, GoPro dataset, YouTube240 dataset and Sony dataset. Quantitative and qualitative results indicate that our model outperforms existing SOTA methods. Besides, experiments on real-world blurry videos also indicate the good generalization ability of our model. 
\end{abstract}


\begin{table*}[t]
	\centering
	\caption{Technical analysis of exisitg BVFI methods.}
	\resizebox{18.5cm}{2.1cm}{
		\begin{tabular}{|l|ccc|ccc|cc|c|}
			\hline
			Methods & \multicolumn{3}{c|}{\textbf{Interpolation}} & \multicolumn{3}{c|}{\textbf{Temporal information}} & \multicolumn{2}{c|}{\textbf{DeblurNet}}&\multirow{3}{*}{\textbf{Multi-frame VFI?}} \\ 
			\cline{1-9}
			& \multicolumn{1}{c|}{\multirow{2}{*}{Non-motion}} & \multicolumn{2}{c|}{Motion estimation} & \multicolumn{1}{c|}{\multirow{2}{*}{Short-term}} & \multicolumn{1}{c|}{\multirow{2}{*}{Long-term}} & \multirow{2}{*}{Alignment} & \multicolumn{1}{c|}{\multirow{2}{*}{CNN}} & \multirow{2}{*}{Transformer}& \\ \cline{1-1} \cline{3-4}
			& \multicolumn{1}{c|}{} & \multicolumn{1}{c|}{Optical flow} & DConv & \multicolumn{1}{c|}{} & \multicolumn{1}{c|}{} &  & \multicolumn{1}{c|}{} &  &\\ \hline
			TNTT~\cite{TNTT} (CVPR19) & \multicolumn{1}{c|}{} & \multicolumn{1}{c|}{\checkmark} &  & \multicolumn{1}{c|}{\checkmark} & \multicolumn{1}{c|}{} &  & \multicolumn{1}{c|}{\checkmark} &  &\\ \hline
			UTI-VFI~\cite{NIPS20} (NIPS20) & \multicolumn{1}{c|}{} & \multicolumn{1}{c|}{\checkmark} &  & \multicolumn{1}{c|}{\checkmark} & \multicolumn{1}{c|}{} &  & \multicolumn{1}{c|}{\checkmark} &  &\checkmark\\ \hline
			BIN~\cite{BIN} (CVPR20) & \multicolumn{1}{c|}{\checkmark} & \multicolumn{1}{c|}{} &  & \multicolumn{1}{c|}{} & \multicolumn{1}{c|}{\checkmark} &  & \multicolumn{1}{c|}{\checkmark} &  &\\ \hline
			ALANET\cite{ALANET} (MM20) & \multicolumn{1}{c|}{\checkmark} & \multicolumn{1}{c|}{} &  & \multicolumn{1}{c|}{} & \multicolumn{1}{c|}{\checkmark} &  & \multicolumn{1}{c|}{\checkmark} &  &\\ \hline
			PRF~\cite{PRF} (TIP21) & \multicolumn{1}{c|}{\checkmark} & \multicolumn{1}{c|}{} &  & \multicolumn{1}{c|}{} & \multicolumn{1}{c|}{\checkmark} &  & \multicolumn{1}{c|}{\checkmark} &  &\\ \hline
			DeMFI~\cite{DeMFI} (ECCV22) & \multicolumn{1}{c|}{} & \multicolumn{1}{c|}{\checkmark} &  & \multicolumn{1}{c|}{\checkmark} & \multicolumn{1}{c|}{} &  & \multicolumn{1}{c|}{\checkmark} &  &\checkmark\\ \hline
			BiT~\cite{BiT} (CVPR23) & \multicolumn{1}{c|}{\checkmark} & \multicolumn{1}{c|}{} &  & \multicolumn{1}{c|}{\checkmark} & \multicolumn{1}{c|}{} &  & \multicolumn{1}{c|}{} & \multicolumn{1}{c|}{\checkmark} &\checkmark\\ \hline
			\textcolor{red}{Ours} & \multicolumn{1}{c|}{} & \multicolumn{1}{c|}{} & \checkmark & \multicolumn{1}{c|}{} & \multicolumn{1}{c|}{\checkmark} & \checkmark & \multicolumn{1}{c|}{} & \checkmark &\checkmark\\ \hline
		\end{tabular}
	}
	\label{tab:BVFI_CP}
\end{table*}
\section{Introduction}
\label{introduction}
Video frame interpolation (VFI)~\cite{DAIN,softmax2020,ABME,VFIFormer} aims to increase the frame rate of a video, which has been widely used in various applications~\cite{usman2016frame}. However, many existing VFI methods assume that the input video frames are free from degradation. In real-world sceneries, the low frame-rate videos are often accompanied by motion blur due to the long exposure time, low shutter frequency, or the movement of the device itself~\cite{BIN}. Therefore, generating high-frame-rate sharp videos from low-frame-rate blurry videos, which we call blurry video frame interpolation (BVFI), is crucial for the practical application of VFI technology.

\begin{figure}
	\centering
	\includegraphics[width=0.9\columnwidth]{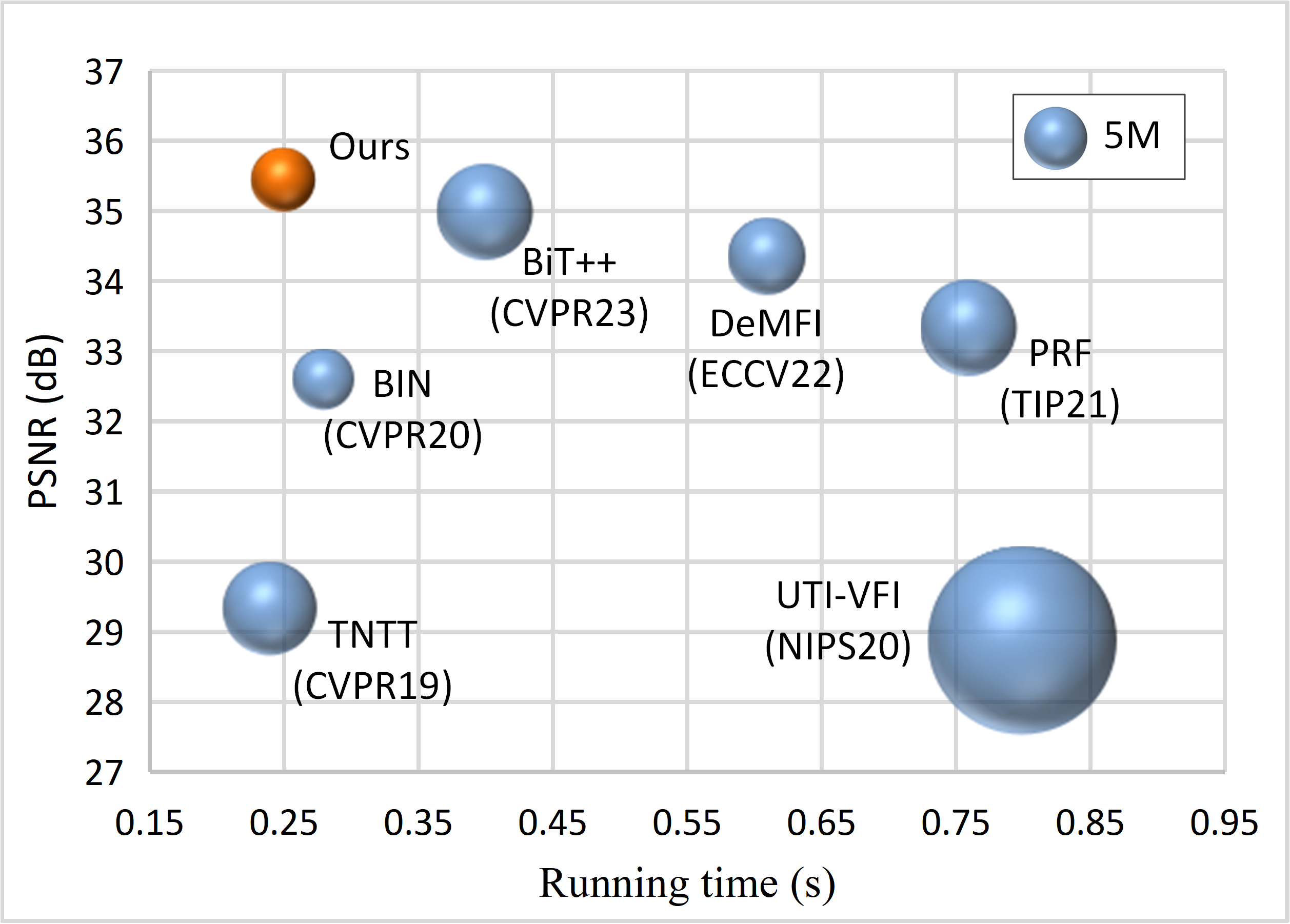} 
	\caption{The model performance, the number of parameters and the running time comparisons of existing SOTA BVFI methods on the Adobe240 testset.}
	\label{fig:plot}
\end{figure}

To solve the BVFI problem, one nature idea is to conduct video frame deblurring~\cite{EDVR,pan2020cascaded}, followed by video frame interpolation. However, it is suboptimal to directly perform the two subtasks in succession. The reason lies mainly in two aspects. First, the pixel error in the deblurring stage may be transferred to the frame interpolation stage, thus influence the interpolation performance~\cite{BIN}. Second, the deblurring process eliminates useful temporal information contained in the blurriness, which increases the difficulty of subsequent VFI tasks.

To handle the BVFI problem more effectively, several methods~\cite{TNTT,NIPS20,BIN,ALANET,PRF,DeMFI, BiT} have been proposed to jointly solve the video deblurring and VFI problems. These methods have demonstrated that the joint solutions are better than simply combining two detached tasks. We analyze these methods from three aspects in detail (see Table~\ref{tab:BVFI_CP}), and draw the conclusion that these methods still fail to fully leverage all valuable information hidden in blurry videos, which leads to suboptimal performance. To fully explore useful information, we propose a simple end-to-end three-stage BVFI framework. Specifically, we separate the BVFI task into three subtasks, i.e., frame interpolation, temporal feature fusion and deblurring. In the following, we will present our detailed analysis from the three aspects as shown in Table~\ref{tab:BVFI_CP}.

\begin{figure}
	\centering
	\includegraphics[width=0.9\columnwidth]{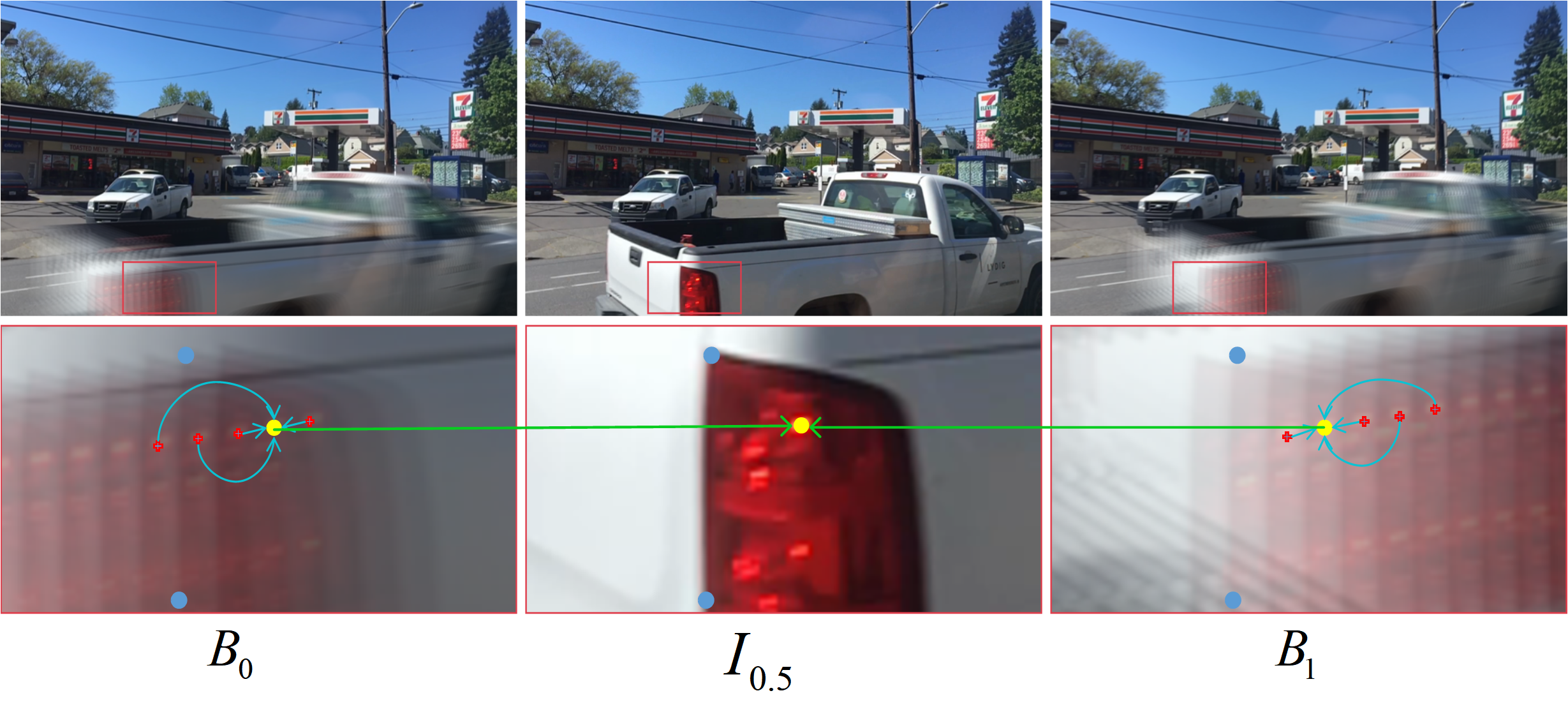} 
	\caption{The sampling process of the deformable convolution for generating an intermediate frame from blurry inputs. The yellow points represent the target pixel that we want to interpolate, the red points represent the sampling locations of the input blurry images, and the blue points are used for locating.}
	\label{fig:sample}
\end{figure}

For the frame interpolation procedure, BIN~\cite{BIN}, ALANet~\cite{ALANET}, PRF~\cite{PRF} and BiT~\cite{BiT} directly use the deep networks to estimate the intermediate frames and do not consider the motion information between consecutive frames. TNTT~\cite{TNTT}, UTI-VFI~\cite{NIPS20} and DeMFI~\cite{DeMFI} take the motion information into consideration, however, they all utilize the optical flows for motion estimation. There are several limitations in using optical flow to solve the BVFI problem. Firstly, the optical flows of blurry frames can be highly uncertain, which makes it difficult to estimate accurate optical flows from such frames. Secondly, optical flow-based methods are limited to single-point sampling, which restricts their ability to fully utilize the rich motion information available in blurry frames. 
Against the drawbacks of optical flow in motion estimation, deformable convolution~\cite{DCNV1,DCNV2} (DConv) has been introduced in various video restoration tasks~\cite{TDAN,EDVR,featureflow,EDSC,Ada_cof}. In fact, DConv can be regarded as a general version of optical flows. To be specific, optical flow-based methods estimate one offset for each pixel position, while DConv predicts multiple offsets for each pixel. The miltiple offsets warping of DConv can collect more diverse information from the input images, thus it should be more robust than single-sampled optical flow-based methods, especially in dealing with motions in blurry videos.
Considering that, we design a temporal deformable network, which can adaptively sample useful information from the blurry frames. Since we consider the temporal information when estimating offsets, our model can interpolate frames at arbitrary time intervals. The benefits of deformable sampling for the BVFI task are shown in Figure~\ref{fig:sample}.

Temporal information is important for video restoration tasks~\cite{wieschollek2017learning,hyun2017online,basicvsr++}. However, existing BVFI methods fail to make full use of useful temporal information. As shown in Table~\ref{tab:BVFI_CP}, TNTT~\cite{TNTT}, UTI-VFI~\cite{NIPS20}, DeMFI~\cite{DeMFI} and BiT~\cite{BiT} only employ the short-term temporal information. Although BIN~\cite{BIN}, ALANet~\cite{ALANET} and PRF~\cite{PRF} utilize long-term temporal information, they ignore the fact that multiple adjacent frames are not aligned with each other. Simply fusing multiple frames may introduce mistakes to the target frame, especially when large motions are involved. To avoid the above problems, we propose a bi-directional recurrent deformable alignment module (Bi-RDAM) to explore the long-term temporal information and avoid the interference of misaligned features.

Another important factor that affects the performance of BVFI is the design of the deblurring network. Existing BVFI methods either use residual network~\cite{DeMFI}, or residual dense networks~\cite{TNTT,NIPS20,BIN}, or UNet~\cite{ALANET} for deblurring, which has limited ability when dealing with high dynamic motion blur. Transformer~\cite{swinir,restormer,liang2022vrt} has advantages in exploring long-range dependencies of an image, which has achieved great success on image and video restoration tasks. Most recently, BiT~\cite{BiT} design a blur interpolation transformer with several multi-scale residual Swin transformer blocks, achieving good performance on BVFI tasks. Based on a powerful transformer model and the Taylor expansion, we design a deep unfolding Taylor approximation network to recursively recover the missing details for the target frames.

\begin{figure*}
	\centering
	\includegraphics[width=2\columnwidth]{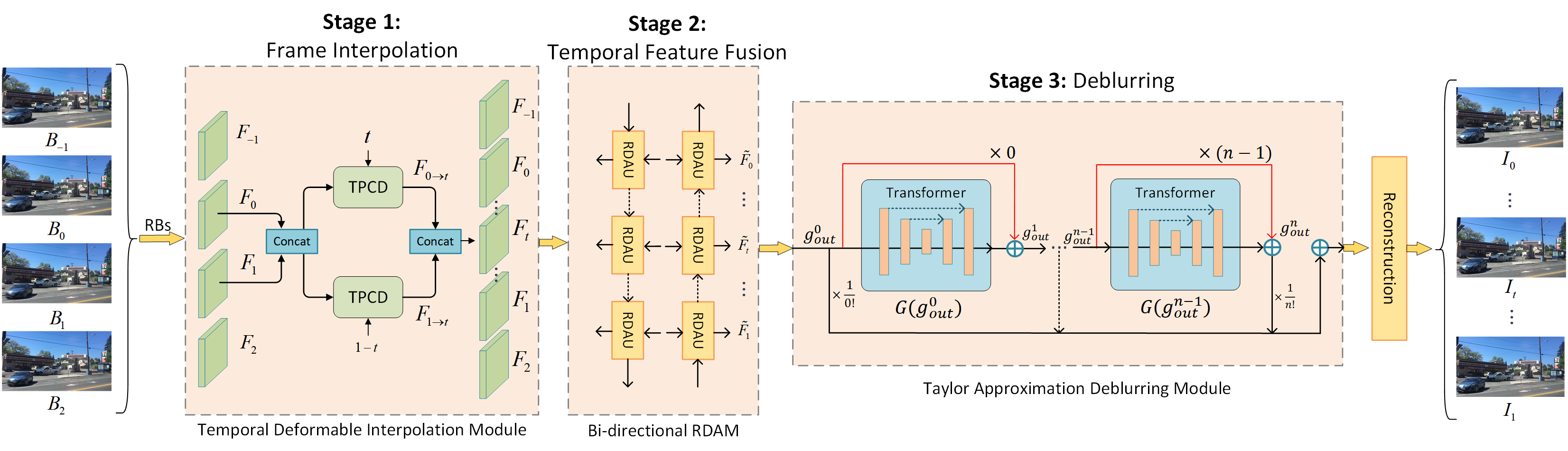} 
	\caption{An overview of the proposed three-stage BVFI framework. It contains three stages: a frame interpolation stage, a temporal feature fusion stage and a deblurring stage. The overall network is trained in an end-to-end manner.}
	\label{fig:overview}
\end{figure*}

Our contributions can be summarized as:
\begin{itemize}
	\item We analyze several crucial technologies that affect BVFI's performance and propose a new end-to-end three-stage BVFI framework with clear task assignment and good expandability to fully explore the abundant useful information in both intra-frame and inter-frame from blurry videos. 
	\item We propose a temporal deformable interpolation module in the frame interpolation stage to adaptively sample useful information from blurry inputs and generate intermediate frames at arbitrary time intervals. 
	\item We design a bi-directional deformable alignment module in temporal feature fusion stage to enable the extraction of long-term temporal information while mitigating the impact of misaligned features. 
	\item We employ a Taylor approximation network for deblurring stage empowered by transformers, enabling the recursive recovery of high-frequency details by leveraging the transformers' ability to capture long-range dependencies.
\end{itemize}

\section{Related work}
\subsection{Video frame interpolation}
Video frame interpolation (VFI) aims to synthesize the intermediate non-existing frames to increase the frame rate of a video sequence. Existing VFI methods can be roughly categorized into optical flow-based methods and kernel-based methods. Optical flow-based methods~\cite{Super2018,DAIN,softmax2020,ABME,IFRNet,VFIFormer} synthesize the intermediate frames by estimate the intermediate optical flows. Niklaus et al. propose kernel-based VFI methods~\cite{Sep_conv,Adapt_Conv}, which applies spatially-adaptive kernels to the input consecutive frames. \cite{Ada_cof} and \cite{EDSC} introduce the deformable convolution to kernel-based methods. They not only estimate the convolution kernels, but also estimate additional offset for each kernel. Most recently, Lei et al.~\cite{FGDCN} propose to use the pre-estimated flow information to guide the learning of the deformable compensation network, which successfully combines the advantages of flow-based with deformable convolution-based VFI methods. Even though these methods have achieved promising results, they are designed for sharp videos. It is challenging for these methods to process blurry videos due to the potential inaccuracy of optical flow/motion estimation.
\subsection{Video deblurring}
Numerous methods have been emerged to solve the video deblurring paroblems. 
The traditional deblurring methods~\cite{bar2007variational,hyun2015generalized,wulff2014modeling} restore sharp frames by jointly estimating the blur kernels and optical flows from the blurry frames. With the development of deep learning, Kim et al.~\cite{hyun2017online} fuse the multiple frame features by designing a recurrent network. Wieschollek et al.~\cite{wieschollek2017learning} design a multi-scale recurrent model, where the features from previous frames can be recurrently transferred to latter frames. Wang et al.~\cite{EDVR} propose a pyramid, cascading and deformable (PCD) alignment module and achieve better alignment performance. Pan et al.~\cite{pan2020cascaded} develop a temporal sharpness prior for video deblurring. Most recently, Zhang et al.~\cite{zhang2022spatio} propose a deformable attention network to fuse the useful information from the blurry images. 
Motivated by the great success of deformable convolution on video deblurring tasks, in this paper, we employ it to directly synthesize the intermediate frames from blurry videos.
\subsection{Joint video deblurring and frame interpolation}
Several methods~\cite{TNTT,NIPS20,ALANET,BIN,PRF,DeMFI} have been proposed to jointly solve the BVFI problem and achieve better performance than directly cascade two separate pre-trained deblurring and VFI networks. 
TNTT~\cite{TNTT} employs two networks to firstly extract sharp frames and then use them to generate intermediate frames. The two networks are jointly optimized in the training phase. ALANET~\cite{ALANET} adaptively fuses features in latent space by designing the network with both self-attention and cross-attention. UTI-VFI~\cite{NIPS20} proposes a general model to solve the BVFI problem without temporal priors. BIN~\cite{BIN} and its larger-sized version PRF~\cite{PRF} design a pyramid network with recurrent ConvLSTM structure to solve both the deblurring and interpolation problems. DeMFI~\cite{DeMFI} considers the optical flow informationn and proposes to interpolate multi-frames from blurry videos. BiT~\cite{BiT} proposes a blur interpolation transformer for solving real-world blurry video deblurring and interpolation.
Although these methods have got promising results, they fail to fully leverage all valuable information from blurry videos. Therefore, the performance of the BVFI model still has a large space for improvement.

\section{Methodology}
\subsection{Framework overview}
Given four input blurry images $B_{-1}$, $B_0$, $B_1$ and $B_2$, we aims to restore sharp frames $I_0$ and $I_1$, and estimate $T$ intermediate frames between $I_0$ and $I_1$. The overall structure of our three-stage BVFI framework is shown in Figure~\ref{fig:overview}. It contains a frame interpolation stage, a temporal feature fusion stage and a deblurring stage. 

For the four input blurry frames, we first employ several residual blocks to map the input frames to the feature domain. In the frame interpolation stage, we propose a temporal PCD (TPCD) module to interpolate an intermediate feature map by considering an additional parameter $t$. $t\in\{0,1\}$ is a time parameter, representing the temporal position that we want to interpolate a new frame. For example, when we want to generate $\times8$ high frame-rate videos, we need to interpolate 7 frames between every two input frames and the temporal position parameter $t\in[1/8,2/8,3/8,4/8,5/8,6/8,7/8]$. In the temporal feature fusion stage, we employ a bi-directional recurrent deformable alignment module (Bi-RDAM) to explore the long-term temporal information for each target frame. In the deblurring stage, a Taylor approximation network is employed to recursively recover the high-frequency details. Finally, a reconstruction layer is used to map the features to the final images.
The detailed structure of the TPCD module, Bi-RDAM module and the Taylor approximation network will be provided in the following section.

\begin{figure}
	\centering
	\includegraphics[width=0.8\columnwidth]{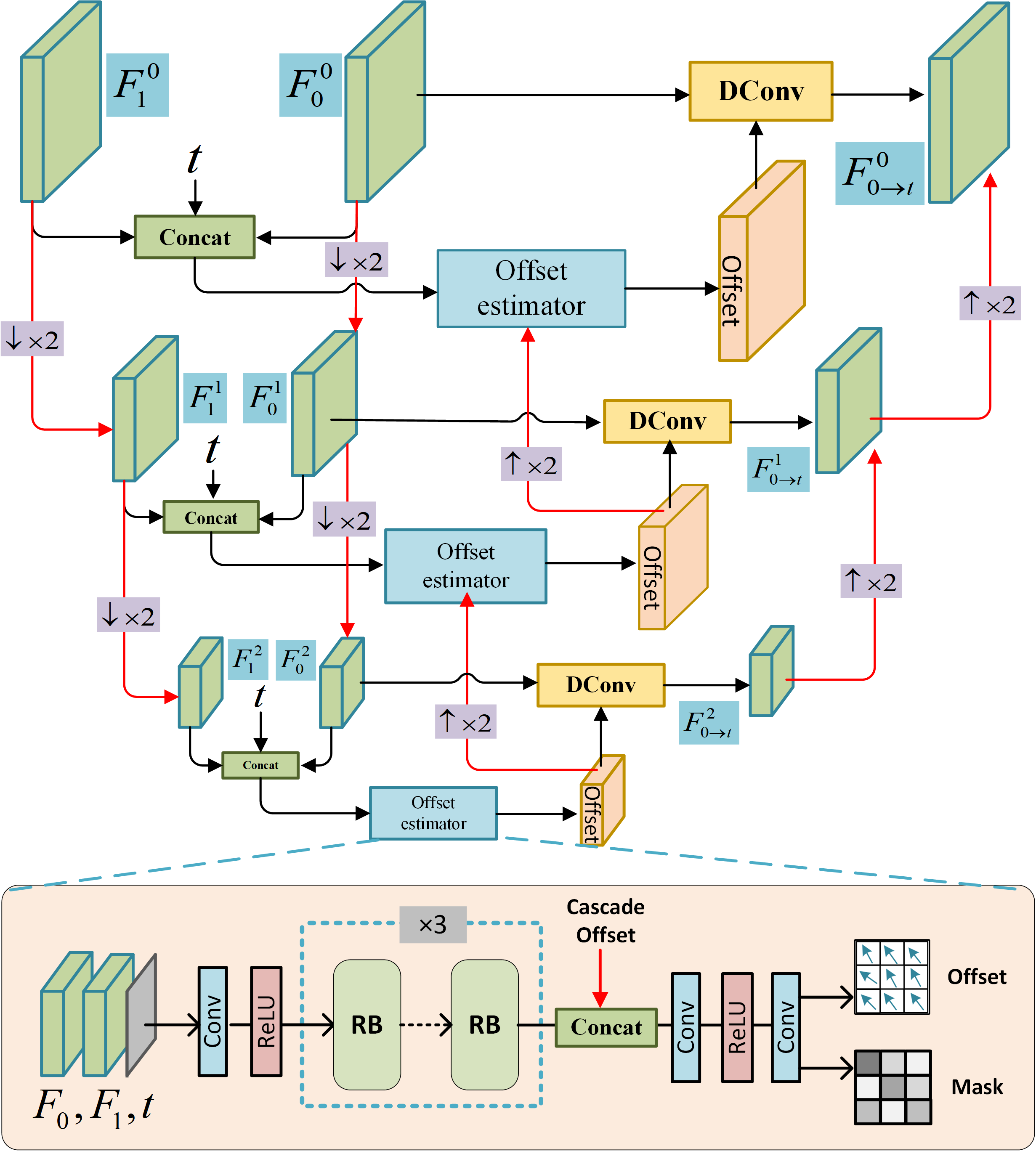} 
	\caption{An overview of the proposed temporal PCD (TPCD) module. RB reprensents residual block.}
	\label{fig:PCD}
\end{figure}
\subsection{Temporal PCD for multi-frame interpoaltion}
The PCD module is first proposed in EDVR~\cite{EDSR}, which is used for frame alignment between adjacent frames for the video super-resolution and video deblurring tasks. 
Motivated by the powerful alignment ability of the PCD module, we propose a temporal PCD (TPCD) module to adaptively sample useful information from the blurry inputs and synthesize the missing frame at an arbitrary time interval. 

The architecture of the proposed TPCD module is shown in Figure~\ref{fig:PCD}. As shown in the figure, the TPCD module takes feature maps $F_0$ and $F_1$ as inputs and synthesizes the intermediate feature $F_{0\rightarrow t}$ by considering an additional temporal position parameter $t$. To be specific, the TPCD module contains three pyramid feature levels. For the $l$-th feature level,  the input features are denoted as $F_0^l$ and $F_1^l$. Firstly, we concatenate the input features $F_0^l$ and $F_1^l$ with the temporal information $t$ and send them to the offset estimator to estimate the deformable offsets and masks. These operations can be formulated as:
\begin{equation}
	\{\Delta p_{0\to t}^l, \Delta m_{0\to t}^l\}=\mathcal{F}_{E}([F_0^l,F_1^l,t],[\Delta p_{0\to t}^{l+1}, \Delta m_{0\to t}^{l+1}]^{\uparrow2}),
\end{equation}
where $\Delta p_{0\to t}^l$ and $\Delta m_{0\to t}^l$ denote the learned offsets and the masks in the $l$-th feature level. $[\Delta p_{0\to t}^{l+1}, \Delta m_{0\to t}^{l+1}]^{\uparrow2}$ is the cascaded offsets from the higher $l+1$ level, $[\cdot]$ is the concatenate operator and $(\cdot)^{\uparrow2}$ represents $\times2$ bilinear interpolation upsampling. $\mathcal{F}_E(\cdot)$ represents the offset estimator network and its detailed network structure is shown at the bottom of Figure~\ref{fig:PCD}. 

Using the learned offsets and masks, we get the feature at temporal position $t$ using deformable convolution and cascaded feature fusion:
\begin{equation}
	F_{0\rightarrow t}^l=\mathcal{F}_{cf}([\mathcal{F}_{dc}(F_0^l,\Delta p_{0\to t}^l,\Delta m_{0\to t}^l),(F_{0\rightarrow t}^{l+1})^{\uparrow2}]),
\end{equation}
where $\mathcal{F}_{cf}(\cdot)$ and $\mathcal{F}_{dc}(\cdot)$ represent the cascaded feature fusion operation and the deformable convolution operation.

After two TPCD modules with two different temporal positions $t$ and $1-t$, we get two aligned feature maps $F_{0\rightarrow t}$ and $F_{1\rightarrow t}$. Thus, the interpolated intermediate feature $F_t$ can be obtained by an adaptive fusion layer:
\begin{equation}
	F_t=\mathcal{F}_{fuse}([F_{0\rightarrow t},F_{1\rightarrow t}]),
\end{equation}
where $\mathcal{F}_{fuse}(\cdot)$ represents the feature fusion layer for synthesizing the intermediate feature at temporal position $t$.
In the same way, we can get as many intermediate features as we want by adjusting the time position parameter $t\in\{0,1\}$.

\begin{figure}
	\centering
	\includegraphics[width=0.9\columnwidth]{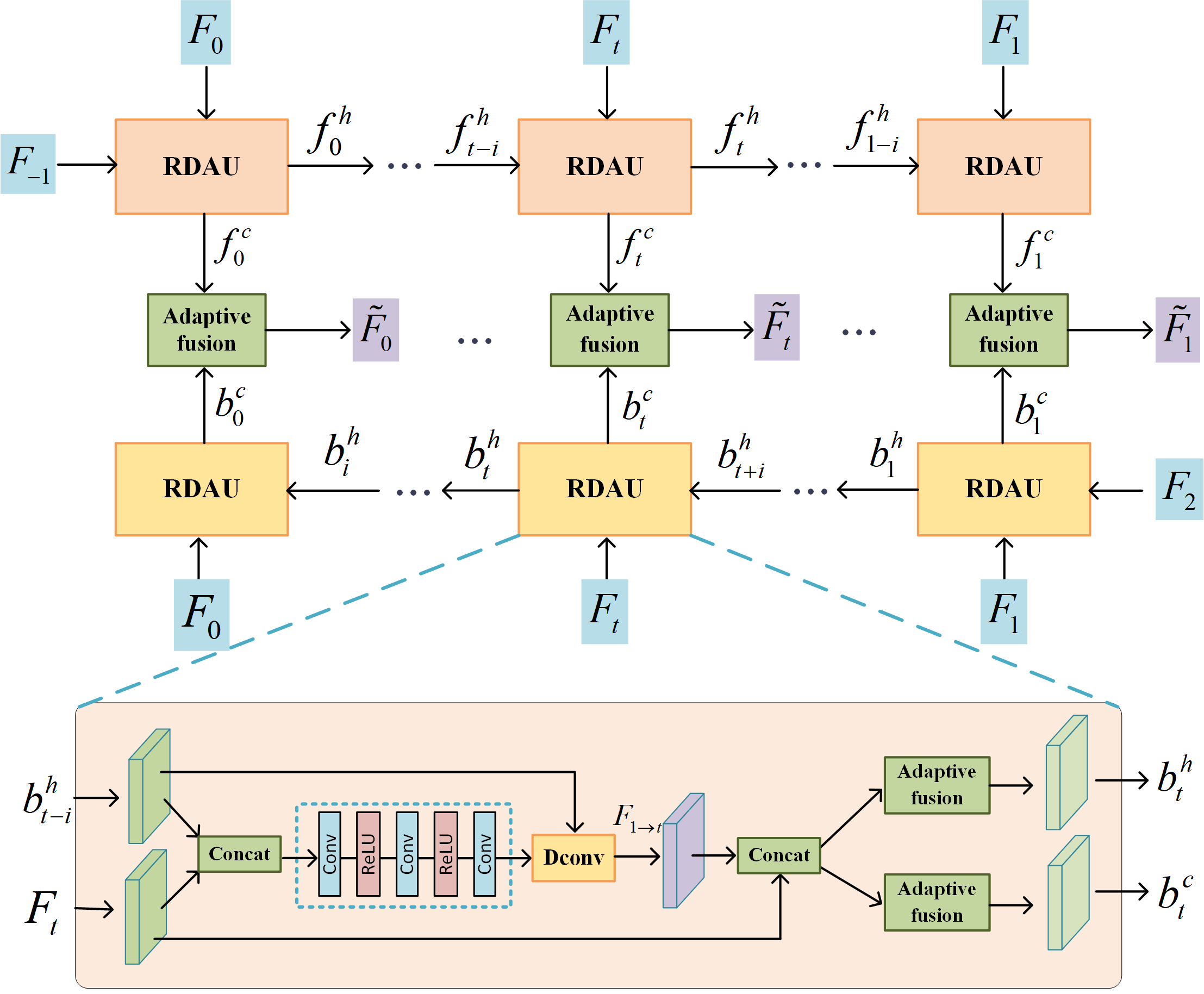} 
	\caption{An overview of the proposed bi-directional recurrent deformable alignment module (Bi-RDAM). $i$ is the time interval of the interpolated frames.}
	\label{fig:RDAM}
\end{figure}
\subsection{Bi-directional recurrent deformable alignment}
Temporal information plays a crucial role in video restoration tasks~\cite{wieschollek2017learning,hyun2017online,basicvsr++}. To fully utilize the temporal information and avoid error accumulation caused by frame misalignment, we propose a bi-directional recurrent deformable alignment module (Bi-RDAM). The overall structure of the proposed Bi-RDAM is shown in Figure~\ref{fig:RDAM}. 
It contains two recurrent branches, a forward branch and a backward branch. The forward branch aims to fuse the information of the current frame with former frames, and the backward branch aims to aggregate the useful information of the current frame with later frames. Each recurrent branch shares the same recurrent deformable alignment unit (RDAU). 

The structure of the RDAU is shown in Figure~\ref{fig:RDAM}. For updating $F_t$ in the backward branch, the input of RDAU contains a current feature $F_t$ and a backward hidden feature $b^h_{t+i}$. Note that $i$ denotes the time interval when interpolation. We first employ a simple deformable convolutional network to align the feature from $b^h_{t+i}$ to $F_t$. Then we use two adaptive fusion modules to generate the backward hidden feature $b_t^h$ and the backward current feature $b_t^c$, respectively. 
When we get the forward current feature $f_t^c$ and the backward current feature $b_t^c$,
the feature with temporal information $\tilde{F}_t$ can be obtained by an adaptive fusion layer. 
These operations can be formulated as:
\begin{equation}
	\left\{\begin{aligned}
		&\{f_t^h,f_t^c\}=\mathcal{F}_{RDAU}(F_t,f_{t-i}^h), \\
		&\{b_t^h,b_t^c\}=\mathcal{F}_{RDAU}(F_t,b_{t+i}^h), \\
		&\tilde{F}_t=\mathcal{F}_{af}([f_t^c,b_t^c]),
	\end{aligned}\right.	
\end{equation}
where $\mathcal{F}_{RDAU}(\cdot)$ denotes the recursive deforamble alignment module, $\mathcal{F}_{af}(\cdot)$ is the adaptive fusion layer, which consists of several $1\times1$ convolutional layers and ReLU layers.

\subsection{Taylor approxiamtion deblurring module}
As listed in Table~\ref{tab:BVFI_CP}, currently available BVFI methods rely on deep convolutional neural networks that are manually designed to remove motion blur. However, these methods have limited ability in modeling long-range dependencies and may lack certain interpretability. Inspired by vision transformer~\cite{liu2021swin} and Taylor expansion, we propose a transformer-empowered deep unfolding Taylor approximation deblurring module to recursively recover the sharp details from the blurry images.

\subsubsection{Taylor approxiamtion formula.}
The deblurring problem can be formulated as an infinite-order Taylor’s series expansion:
\begin{equation}\label{Eq:over}
	\begin{aligned}
		x &= \mathcal{H}(y_0) = \mathcal{H}(y+\epsilon)\\
		&=\mathcal{H}(y) + \frac{1}{1!} \mathcal{H}^{'}(y) \epsilon + \frac{1}{2!}\mathcal{H}^{(2)}(y) (\epsilon)^{2}  + ... + \mathcal{R}_{n}(y_0)
	\end{aligned}
\end{equation}
where $x$ denotes the sharp image, $y_0$ and $y$ are the blurry image and its noise-free version, respectively. $\epsilon$ represent the noise, $\mathcal{H}(\cdot)$ represents the mapping function for deblurring. $\mathcal{R}_{n}(y_0)$ is the Lagrange remainder term.
In actuality, Eq.~(\ref{Eq:over}) can be separated into two parts, i.e., a constant approximation part and a high-order part. In our model, we regard the blurry and the interpolated features as the constant approximation. And here we mainly focus on solving the rest high-order parts. 

Denoting the $k$ order of Eq.~(\ref{Eq:over}) as $\mathcal{H}^{(k)}(y)(\epsilon)^k$. Differentiating it for $y$, we can get the $k+1$ order as 
\begin{equation}
	\big(\mathcal{H}^{(k)}(y)(\epsilon)^k\big)^{'} \times \epsilon = \mathcal{H}^{(k+1)}(y)(\epsilon)^{k+1} -k\mathcal{H}^{(k)}(y)(\epsilon)^k.
	\label{Eq:expand}
\end{equation}

We further denote the $k$ order $\mathcal{H}^{(k)}(y)(\epsilon)^k$ as $g_{out}^{k}$. Here we employ a network, named $\mathcal{G}(\cdot)$ to solve its $k+1$ order $g_{out}^{k+1}$. Referring Eq.~(\ref{Eq:expand}), the connection between the output of $k$ order and $k+1$ order can be formulated as:
\begin{equation}
	\label{iteration}
	g_{out}^{k+1} = \mathcal{G}(g_{out}^{k}) + kg_{out}^{k}.
\end{equation}

Based on Eq.~(\ref{iteration}), we design our deep-unfolding Taylor approximation deblurring network as shown in Figure~\ref{fig:overview}. In this model, each derivative is implemented as a recursion of a transformer network, which is used to recursively recover the high-frequency details by leveraging the transformers' ability to capture long-range dependencies. Similar idea can be seen in~\cite{fu2021unfolding}.

\begin{figure}
	\centering
	\includegraphics[width=1\columnwidth]{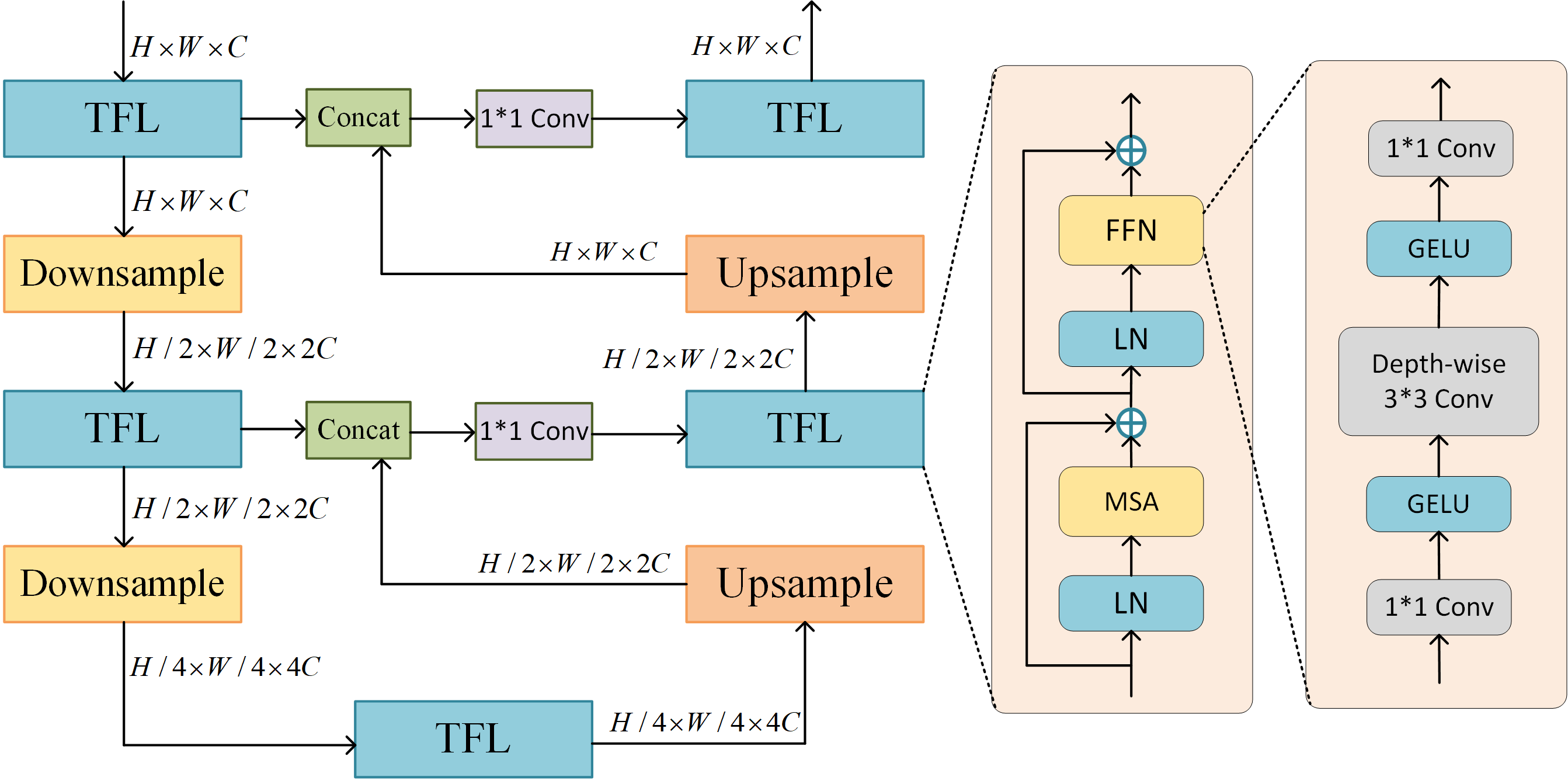} 
	\caption{An overview of the transformer model.}
	\label{fig:FT}
\end{figure}

\subsubsection{Transformer model for one recursion}
Inspired by~\cite{liu2021swin,swinir}, we design a lightweight transformer to recover detailed information from blurry features. 
Its overall structure is shown in Figure~\ref{fig:FT}, which presents a U-shaped structure. The core component is the transformer layer (TFL) which consists of two layer norm (LN) operators, a multi-head self-attention (MSA) and a feedforward network (FFN).

\textbf{MSA.} 
Denoting $X \in \mathbb{R}^{H\times W \times C}$ as the input tokens. Then $X$ can be linearly projected into query: $Q\in \mathbb{R}^{H\times W\times C}$, key: $K\in \mathbb{R}^{H\times W\times C}$ 
and value: $V\in \mathbb{R}^{H\times W\times C}$ as 
\begin{equation}
	Q = X W^{Q}, K = X W^K, V = X W^V,
\end{equation} 
where $W^Q, W^K, W^V\in\mathbb{R}^{C\times C}$ are learnable parameters.
$Q, K, V$ are divided into non overlapping windows with the size of $M\times M$, and then they are reshaped into $\mathbb{R}^{\frac{HW}{M^2}\times M^2 \times C}$.
Subsequently, $Q, K, V$ are splitted into $h$ heads: $Q = [Q^1, ... Q^h], K = [K^1, ..., K^h]$ and $V = [V^1, ..., V^h]$. Each head has the dimension of $d_h = \frac{C}{h}$. The self-attention $A^i \in \mathbb{R}^{\frac{HW}{M^2} \times M^2 \times d_h}$ is calculated inside each head as
\begin{equation}
	A^{i} = softmax\bigg(\frac{Q^{i}K{^{i}}^T }{\sqrt{d_h}}  + P^{i} \bigg)V^{i},   i = 1, ..., h,
\end{equation}
where $P^{i} \in \mathbb{R}^{M^2\times M^2}$ denotes the learnable parameters with corresponding position information.
Then the outputs can be obtained by a linear projection as 
\begin{equation}
	MSA(X) = \sum_{i=1}^{h} A^i W^i. \label{Eq:F-msa}
\end{equation}
$W^i \in \mathbb{R}^{d_h\times C}$ are the learnable parameters. Finally, we reshape the result and get the final output of MSA $X_{out} \in \mathbb{R}^{H\times W\times C}$.

\begin{table*}[t]
	\centering
	\caption{Quantitative results on the Adobe240~\cite{su2017deep}, GoPro240~\cite{nah2017deep} and YouTube240~\cite{DeMFI} test sets for deblurring and single-frame interpolation ($\times2$)). The \textbf{bolded} represents the best, and the \underline{underlined} represents the second-best results. "++" denotes the model uses the temporally symmetric ensembling strategy.}
	\resizebox{18.5cm}{1.6cm}{
		\begin{tabular}{l|c|c|cccccc|cccccc|cccccc}
			\toprule
			\multirow{3}{*}{Method} & \multirow{3}{*}{\begin{tabular}[c]{@{}c@{}}Runtime\\			
					(seconds)\end{tabular}} & \multirow{3}{*}{\begin{tabular}[c]{@{}c@{}}Params\\ (million)\end{tabular}} & \multicolumn{6}{c}{Deblurring}                                                                            & \multicolumn{6}{c}{Interpolation(x2)}                                                                     & \multicolumn{6}{c}{Comprehensiveness}                                                                     \\
			\cline{4-21}
			&&& \multicolumn{2}{c}{Adobe240}                        & \multicolumn{2}{c}{GoPro240} &\multicolumn{2}{c}{YouTube240}                       & \multicolumn{2}{c}{Adobe240}                        & \multicolumn{2}{c}{GoPro240} &\multicolumn{2}{c}{YouTube240}                       & \multicolumn{2}{c}{Adobe240}                        & \multicolumn{2}{c}{GoPro240}   &\multicolumn{2}{c}{YouTube240}                     \\
			\cline{4-21}
			& & & PSNR & SSIM & PSNR & SSIM & PSNR & SSIM & PSNR & SSIM & PSNR & SSIM & PSNR & SSIM & PSNR & SSIM & PSNR &SSIM & PSNR & SSIM \\
			\hline
			UTI-VFI~\cite{NIPS20}                                     & 0.80                                                                                             & 43.4                                                                                                & 28.73                    & 0.8656                   & 27.78                    & 0.8612  &-&-                 & 29.00                    & 0.8690                   & 29.79                    & 0.8700 &-&-                  & 28.87                    & 0.9673                   & 28.78                    & 0.8655 &-&-                  \\
			TNTT~\cite{TNTT}                                        & 0.24                                                                                             & 10.8                                                                                                & 29.40                    & 0.8734                   & 28.44                    & 0.9107 &-&-                   & 29.24                    & 0.8754                   & 27.84                    & 0.8928  &-&-                 & 29.32                    & 0.8744                   & 28.84                    & 0.9010 &-&-                  \\
			BIN~\cite{BIN}                                         & 0.28                                                                                             & 4.68                                                                                              & 32.67                    & 0.9236                   & 30.66                         & 0.8956        &32.50&0.9257                  & 32.51                    & 0.9280                   & 30.98                         & 0.9055 &32.07&0.9162                        & 32.59                    & 0.9258                   &  30.82                        & 0.9006    &32.29&0.9210                      \\
			PRF~\cite{PRF}                                        & 0.76                                                                                             & 11.4                                                                                                & 33.33                    & 0.9319                   & 31.05                    & 0.9064 &32.70&0.9282                  & 33.31                    & 0.9372                   & 31.06                    & 0.9070  &32.36&0.9199                 & 33.32                    & 0.9346                   & 31.06                    & 0.9067 &32.53& 0.9241                 \\
			ALANET~\cite{ALANET}                                     & -                                                                                                & -                                                                                                   & 33.71                    & 0.9329                   & -                         & -    &-&-                     & 32.98                    & 0.9362                   & -                         & -  &-&-                       & 33.34                    & 0.9355                   &   -                       &  -    &-&-                    \\
			DeMFI~\cite{DeMFI}                                  & 0.61                                                                                             & 7.41                                                                                                & \underline{34.19}                    & \underline{0.9410}                   & \underline{30.82}                        & \underline{0.8991}       &\underline{33.52}&\underline{0.9310}                 & \underline{34.49}                    & \underline{0.9486}                   & \underline{31.53}                        &  \underline{0.9165}        &\underline{33.19}&\underline{0.9270}               & 34.34                   & 0.9448                  &  \underline{31.18}                       &  \underline{0.9078}     &\underline{33.36}& \underline{0.9290}                 \\
			BiT~\cite{BiT}                                     & 0.20                                                                                               & 11.3                                                                                                  &                    - & -                   & -                         & -    &-&-                     & -                    & -                   & -                         & -  &-&-                       & 34.34                    & 0.9480                   &   -                       &  -    &-&-                    \\
			BiT++~\cite{BiT}                                     & 0.40                                                                                               & 11.3                                                                                                  &                    - & -                   & -                         & -    &-&-                     & -                    & -                   & -                         & -  &-&-                       & \underline{34.97}                    & \underline{0.9540}                   &   -                       &  -    &-&-                    \\
			Ours     & 0.25 &5.04                                                                                            &\textbf{35.24}                &\textbf{0.9527}                           & \textbf{31.89}                         & \textbf{0.9201}     &\textbf{33.55}&\textbf{0.9325}                    &  \textbf{35.62}               & \textbf{0.9584}                      &\textbf{32.66}                          &  \textbf{0.9338 }      &\textbf{33.96}&\textbf{0.9386}                 & \textbf{35.43}                     & \textbf{0.9556}                         & \textbf{32.28}                         &  \textbf{0.9270} &\textbf{33.76}& \textbf{0.9356}                     
			\\\bottomrule
		\end{tabular}
	}
	\label{tab:BVFIx2}
\end{table*}

\begin{table*}[]
	\centering
	\caption{Quantitative results on the Adobe240~\cite{su2017deep}, GoPro240~\cite{nah2017deep} and YouTube240~\cite{DeMFI} test sets for deblurring and multi-frame interpolation ($\times8$)). The \textbf{bolded} represents the best, and the \underline{underlined} represents the second-best results.}
	\resizebox{18.5cm}{1.2cm}{
		\begin{tabular}{l|c|c|cccccc|cccccc|cccccc}
			\toprule
			\multirow{3}{*}{Method} & \multirow{3}{*}{\begin{tabular}[c]{@{}c@{}}Runtime\\ (seconds)\end{tabular}} & \multirow{3}{*}{\begin{tabular}[c]{@{}c@{}}Params\\ (million)\end{tabular}} & \multicolumn{6}{c}{Deblurring}                                                                                                             & \multicolumn{6}{c}{Interpolation(x8)}                                                                                                      & \multicolumn{6}{c}{Comprehensiveness}                                                                                                      \\
			\cline{4-21}
			&   &    & \multicolumn{2}{c}{Adobe240}                        & \multicolumn{2}{c}{GoPro240}                        & \multicolumn{2}{c}{Youtube240} & \multicolumn{2}{c}{Adobe240}                        & \multicolumn{2}{c}{GoPro240}                        & \multicolumn{2}{c}{Youtube240} & \multicolumn{2}{c}{Adobe240}                        & \multicolumn{2}{c}{GoPro240}                        & \multicolumn{2}{c}{Youtube240} \\
			\cline{4-21}
			&  &   & PSNR & SSIM & PSNR & SSIM & PSNR          & SSIM           & PSNR & SSIM & PSNR & SSIM & PSNR          & SSIM           & PSNR & SSIM & PSNR & SSIM & PSNR          & SSIM           \\\hline
			UTI-VFI\cite{NIPS20}                                    & 0.80                                                                                             & 43.4                                                                                                & 28.73                    & 0.8656                   & 25.66                    & 0.8085                   & 28.61         & 0.8891         & 28.66                    & 0.8648                   & 25.63                    & 0.8148                   & 28.64         & 0.8900         & 28.87                    & 0.9673                   & 28.78                    & 0.8655                   & 28.64         & 0.8899         \\
			TNTT\cite{TNTT}                                       & 0.24                                                                                             & 10.8                                                                                                & 29.40                    & 0.8734                   & 26.48                    & 0.8085                   & 29.59         & 0.8891         & 29.45                    & 0.8765                   & 26.68                    & 0.8148                   & 29.77         & 0.8901         & 29.32                    & 0.8744                   & 28.84                    & 0.9010                   & 29.75         & 0.8899         \\
			PRF\cite{PRF}                                         & 0.76                                                                                             & 11.4                                                                                                & 33.33                    & 0.9319                   & 30.27                    & 0.8866                   & 32.37         & 0.9199         & 28.99                    & 0.8774                   & 25.68                    & 0.8053                   & 29.11         & 0.8919         & 33.32                    & 0.9346                   & 30.82                    & 0.9006                   & 29.52         & 0.8954         \\
			DeMFI\cite{DeMFI}                                  & 0.61                                                                                             & 7.41                                                                                                & \underline{34.19}                    & \underline{0.9410}                  & \underline{30.82}                    & \underline{0.8991}                   & \underline{33.31}         & \underline{0.9282}         & \underline{34.29}                    & \underline{0.9454}                   & \underline{31.25}                    & \underline{0.9102}                   & \underline{33.33}         & \underline{0.9300}         & \underline{34.28}                    & \underline{0.9449}                  & \underline{31.20}                    & \underline{0.9088}                   & \underline{33.33}         & \underline{0.9298}         \\
			Ours & 0.25 & 5.04  & \textbf{35.04}              &\textbf{0.9513}          &    \textbf{31.66}    & \textbf{0.9166} &\textbf{33.36} &\textbf{0.9312}         & \textbf{35.55}      &  \textbf{0.9569}        & \textbf{32.47}              &  \textbf{0.9296}&
			\textbf{33.78}&\textbf{0.9356}       &\textbf{35.49}        &  \textbf{0.9562}       & \textbf{32.37}      & \textbf{ 0.9280}&\textbf{33.73}&\textbf{0.9351}
			\\ \bottomrule          
		\end{tabular}
	}
	\label{tab:x8}
\end{table*}
\section{Experiments}
\subsection{Datasets}
\textbf{Adobe240 Dataset.} Adobe240 dataset~\cite{su2017deep} is used to train our model. It contains 120 videos at 240fps. Each frame has a resolution of $1280\times720$. In the training phase, we select 112 videos to construct the training set and the remaining 8 videos for evaluation. Following~\cite{BIN,ALANET,DeMFI}, we synthesize the blurry frames by averaging 11 consecutive claer frames. The stride of this procedure is set to 8. In this way, we get the synthesized blurry image with a long exposure time. The generation blurry videos are 30fps and they are downsized to $640\times352$ as done in~\cite{BIN,ALANET,DeMFI}.

\textbf{GoPro240 Dataset.} The GoPro240 dataset~\cite{nah2017deep} contains 33 high-quality videos. The original sharp videos have a frame rate of 240 fps. Each frame in this dataset has a resolution of $1280\times720$.  Following~\cite{PRF,DeMFI}, We use 11 of those videos to evaluate our model.

\textbf{YouTube240 Dataset.} We emply the YouTube240 Dataset provided by~\cite{DeMFI} to test our model. It contains 60 YouTube videos with a resolution of $1280\times720$ at 240fps. This test set contains diverse scenes captured by different devices. Following~\cite{BIN,ALANET,DeMFI}, we also resized them to $640\times352$ when testing.

\textbf{Sony Dataset.} It [4] has 40 videos at 250fps, which are captured by a Sony RX V camera. To avoid domain bias to the different capturing devices, similar to \cite{PRF}, we also use this dataset to fine-tune the model pre-trained on the Adobe240 dataset. We select the first 35 videos in Sony dataset for network training and the rest 5 videos for testing. Following~\cite{TNTT, PRF}, we also employ the real blurry videos with 25fps to evaluate our model on real-world sceneries.

\subsection{Model implementation details}
Our model is realized in PyTorch using two NVIDIA RTX3090 GPUs. In the trainging phase, the batch size is set to 2 and the patch size is set to $192\times192$. Samples are augmented by random rotation and flipping. We adopt Adam optimizer~\cite{kingma2014adam} to optimize the propose model and the learning rate is decayed using the cosine annealing scheme~\cite{loshchilov2016sgdr}. The initial learning rate is set to $1\times10^{-4}$ and decayed to $1\times10^{-5}$. The model totally iterates 600K, about 100 epochs. Charbonnier loss~\cite{charbonnier1994two} is used to supervise all the reconstructed frames and the overall network is trained in an end-to-end manner. 

\subsection{Evaluation Metrics}
We employ Peak signal-to-noise ratio (PSNR), structural similarity (SSIM) and motion smoothness (MS)~\cite{BIN,PRF} to evaluate our model. The higher PSNR and SSIM indicate better performance. The lower MS indicates better results. 

\begin{figure*}[t]
	\centering
	\includegraphics[width=1.9\columnwidth]{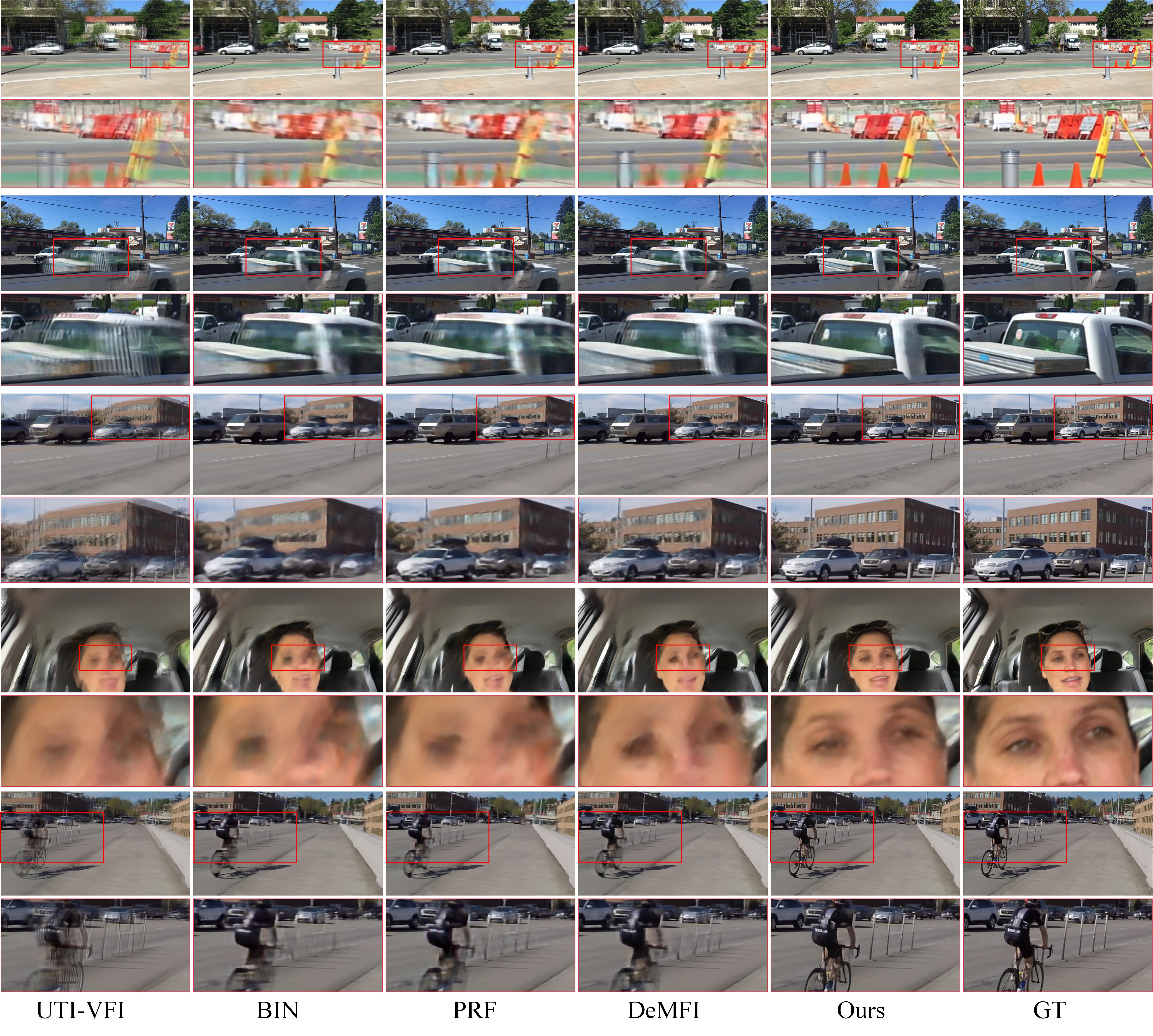} 
	\caption{Visual comparisons of our method with existing SOTA methods (UTI-VFI~\cite{NIPS20}, BIN~\cite{BIN}, PRF~\cite{PRF} and DeMFI~\cite{DeMFI}) on Adobe240~\cite{su2017deep} test set for $\times2$ BVFI.}
	\label{fig:x2}
\end{figure*}
\begin{figure*}[t]
	\centering
	\includegraphics[width=2\columnwidth]{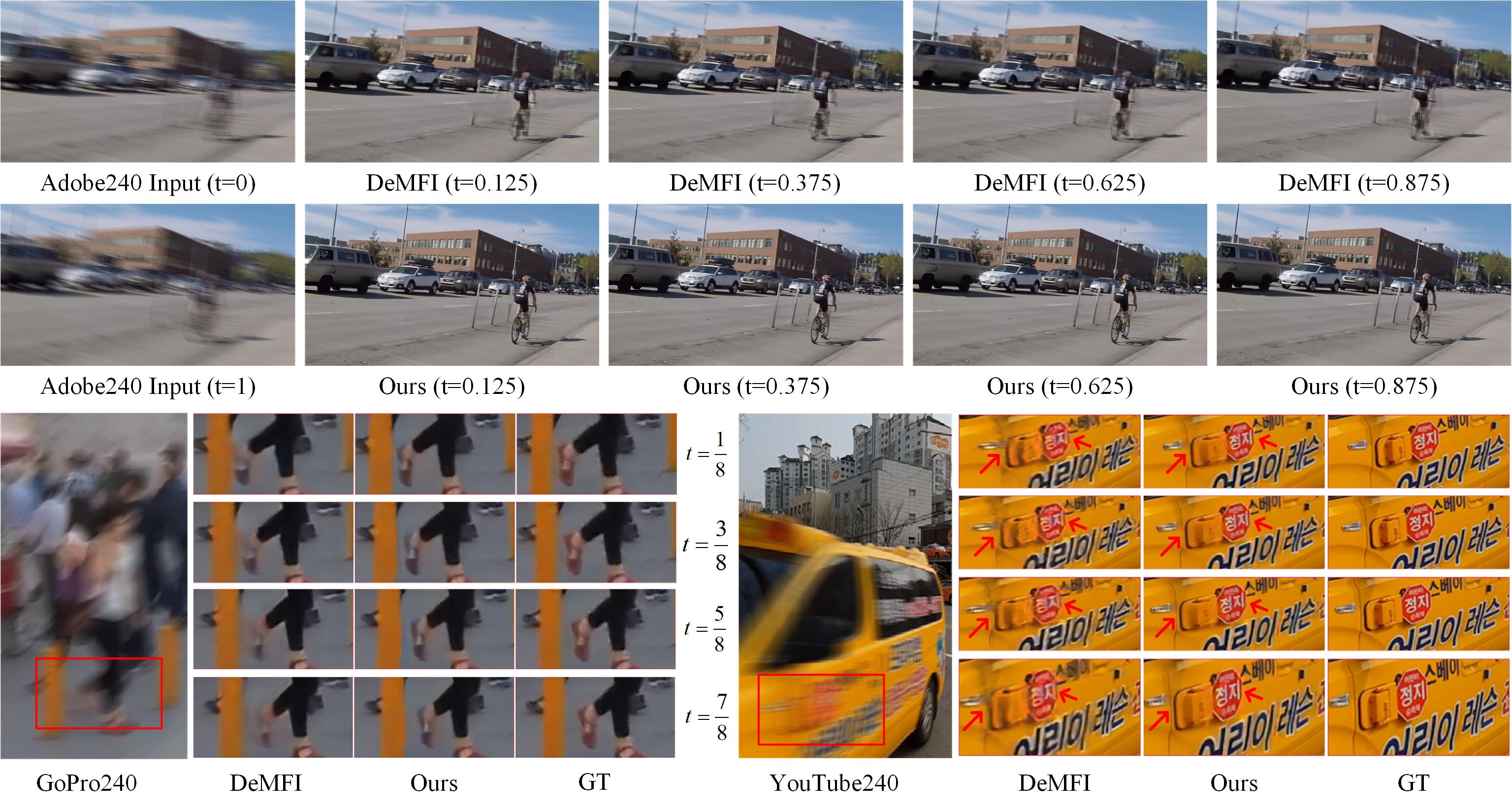} 
	\caption{Visual comparisons of our method with DeMFI~\cite{DeMFI} on Adobe240~\cite{su2017deep}, GoPro240~\cite{nah2017deep} and YouTube240~\cite{DeMFI} testset for $\times8$ BVFI.}
	\label{fig:x8}
\end{figure*}

\begin{figure}[t]
	\centering
	\includegraphics[width=1\columnwidth]{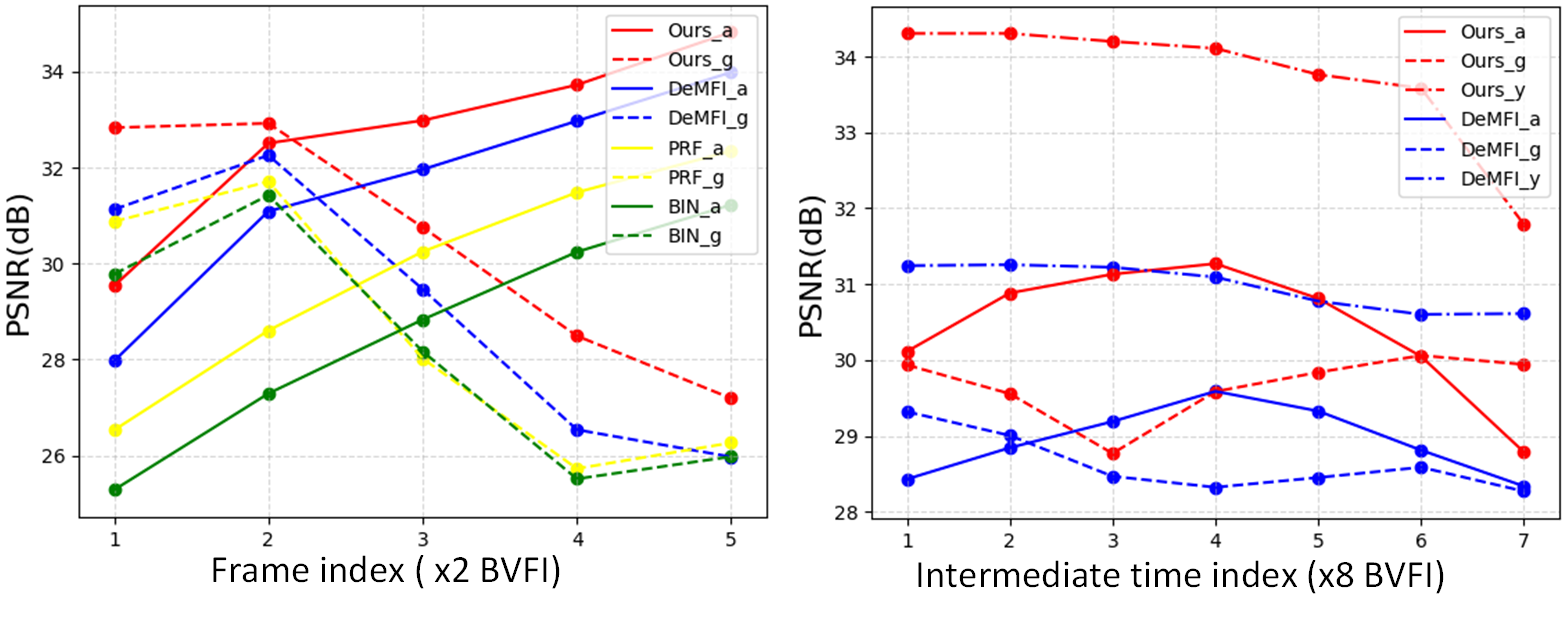} 
	\caption{The PSNR curves of several consecutive frames generated by different methods. “-a”, “-g” and “-y” denote the Adobe240~\cite{su2017deep}, GoPro240~\cite{nah2017deep} and YouTube240~\cite{DeMFI} test sets, respectively.}
	\label{fig:curve}
\end{figure} 

\begin{figure}[t]
	\centering
	\includegraphics[width=0.8\columnwidth]{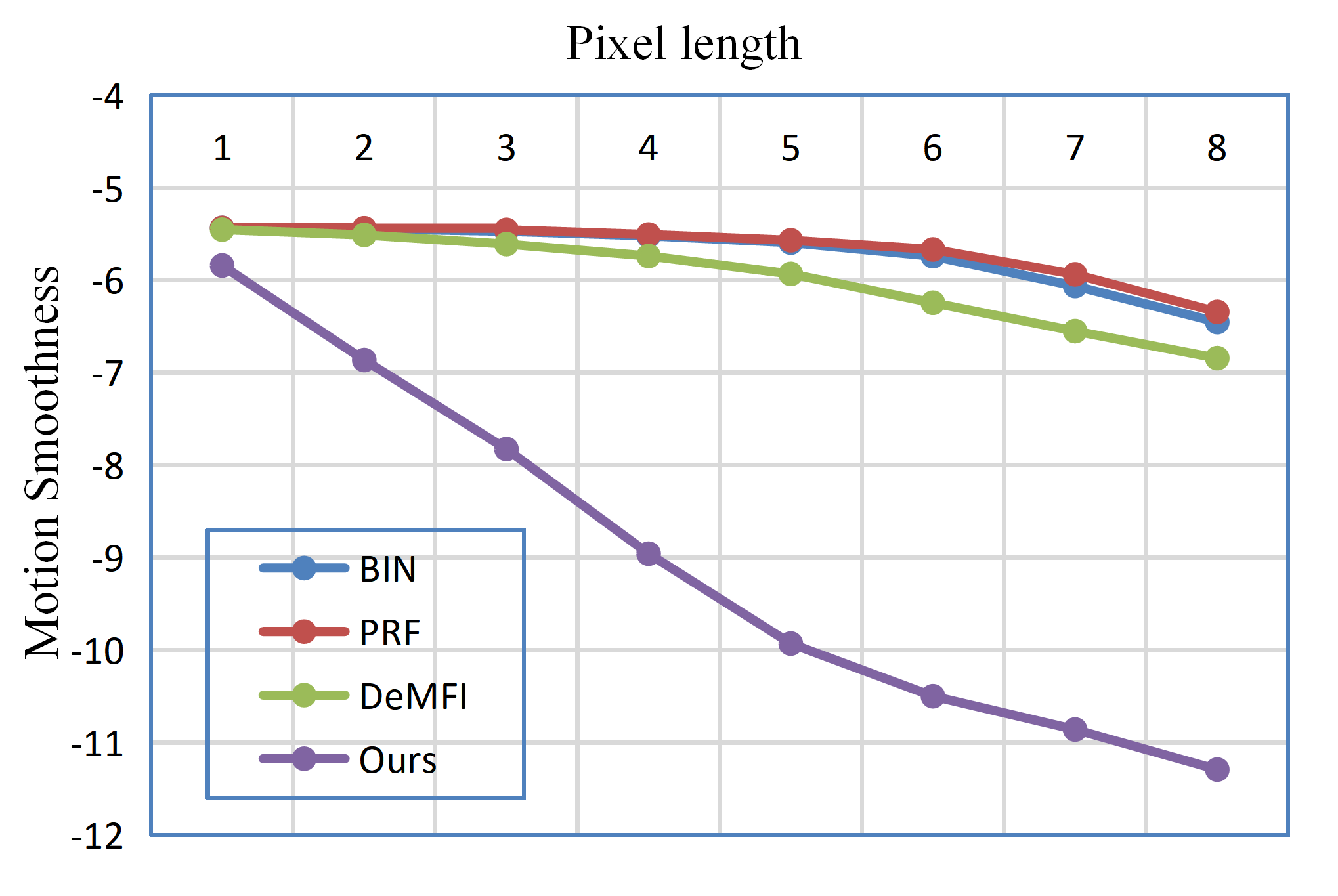} 
	\caption{Motion smoothness~\cite{BIN,PRF} comparisons of existing SOTA methods on the Adobe240 test set. The lower motion smoothness represents better results.}
	\label{fig:MS}
\end{figure}

\subsection{Comparison to SOTA Methods}
\subsubsection{Quantitative Comparison}
We compare our method with six previous SOTA BVFI methods, including TNTT~\cite{TNTT}, UTI-VFI~\cite{NIPS20}, BIN~\cite{BIN}, PRF~\cite{PRF}, ALANET~\cite{ALANET}, DeMFI~\cite{DeMFI} and BiT~\cite{BiT}. All of these comparison methods jointly optimize the deblurring and VFI problems. Note that the TNTT, BIN, PRF and ALANET are designed for $\times2$ interpolation.  Although we can generate multiple frames by recursively passing the interpolation model, it may propagate errors to the later interpolation frames. 
UTI-VFI, DeMFI and our model can realize arbitrary multi-frame interpolation. 
Here we compare the performance of these methods on $\times2$ and $\times8$ BVFI tasks.

\textbf{Deblurring and $\times2$ interpolation:} Table~\ref{tab:BVFIx2} shows the quantitative results of our model with existing SOTA methods on the Adobe240~\cite{su2017deep}, GoPro240~\cite{nah2017deep} and YouTube~\cite{DeMFI} test sets for the $\times2$ BVFI task. Our method gets the best performance on the three test sets. Specifically, our model achieves 1.09dB, 1.10dB and 0.40dB gains against DeMFI~\cite{DeMFI} on the three test sets. Compared with BiT++, our model also outperforms it by 0.46dB PSNR on the Adobe240 test set. It's worth noting that BiT++ employs the temporally symmetric ensembling strategy~\cite{BiT} to improve the model performance.  

\textbf{Deblurring and $\times8$ interpolation.} Table~\ref{tab:x8} compares the quantitative results of our model with existing SOTA methods on the Adobe240~\cite{su2017deep}, GoPro240~\cite{nah2017deep} and YouTube240~\cite{DeMFI} test sets for the $\times8$ BVFI task. Compared to second place DeMFI~\cite{DeMFI}, our method outperforms it by 1.21dB, 1.17dB and 0.4dB on the three benchmark test sets. Our model gets a comparable deblurring results with DeMFI on the YouTube240 testset, but the interpolation performance is 0.77dB higher than it. In Figure~\ref{fig:curve}, we randomly select several consecutive frames from the reconstruction videos of different methods to visually show the PSNR changes over time, our method outperforms existing SOTA methods significantly. 

\textbf{Efficiency analysis:} For a fair comparison, we also test the running time of our model for $640\times352$-sized frames on an RTX2080Ti GPU~\cite{BIN,PRF}. As listed in Table~\ref{tab:BVFIx2}, our model also has advantages in terms of the running speed and the number of parameters compared with existing SOTA methods. In Figure~\ref{fig:plot}, we visually compare the model performance, the running time and the number of parameters of our method with existing SOTA BVFI methods on the Adobe240 testset. Our approach has comprehensive advantages compared with existing SOTA methods.

\textbf{Motion smooth evaluation:} In terms of the motion smoothness evaluation, following~\cite{BIN,PRF}, we first calculate the differential optical flows between three input frames and three corresponding reference frames. Then we compute the motion smoothness (MS) of three frames by considering the pixel error $l$, where $l\in[1,8]$. The lower MS represents better results. To compare the motion smoothness of our method and existing SOTA methods, we randomly select a video sequence from the Adobe240 testset and calculate their average motion smoothness. In Fig.~\ref{fig:MS}, we show their MS index by considering different pixel error lengths. Note that lower MS indicates better results. Our approach has advantages in motion smoothness.

\subsubsection{Qualitative Comparison}
In figure~\ref{fig:x2}, we visualize the reconstruction results of different methods on the Adobe240 test set for $\times2$ BVFI task. From the figure, we can clearly see that our method can restore sharp boundaries of the moving objects. In figure~\ref{fig:x8}, we compare the results of our method with DeMFI~\cite{DeMFI} on the GoPro240 and YouTube240 test sets for $\times8$ BVFI task. Our method can accurately predict continuous motions and restore more detailed textures.

\begin{figure}[t]
	\centering
	\includegraphics[width=1\columnwidth]{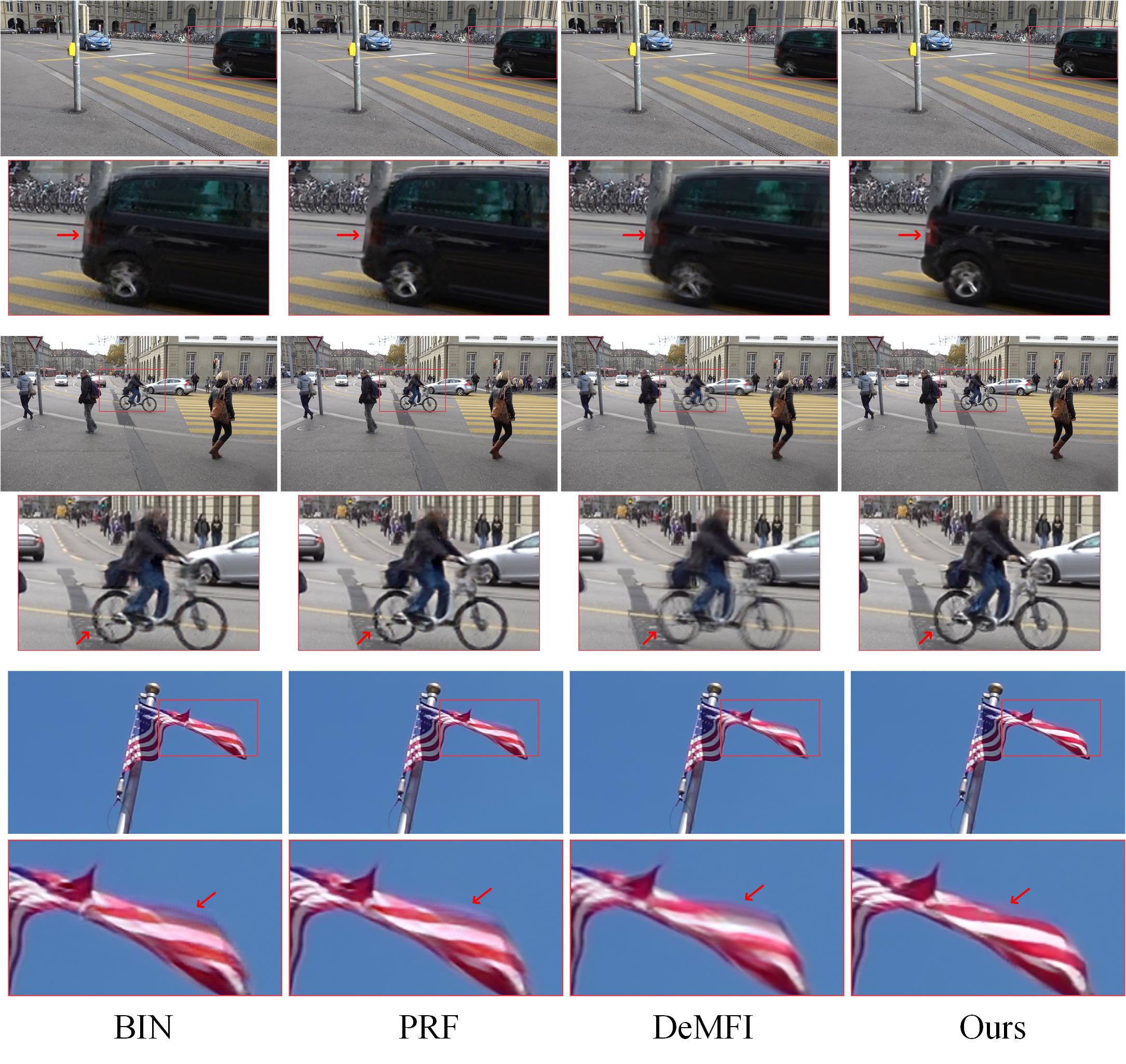} 
	\caption{Visual comparisons of our model with existing SOTA methods on real-world blurry videos capturing from the Sony camera~\cite{TNTT}.}
	\label{fig:real}
\end{figure}
\subsubsection{Evaluation on real-world blurry videos}
We also test our model on real-world blurry videos captured by a Sony camera~\cite{TNTT}. To avoid the domain gap from different capturing devices, we also employ the high framerate videos of work~\cite{TNTT} to fine-tune the model pre-trained on Adobe data set. The detailed setting of the fine-tuning are the same as~\cite{PRF}. For a fair comparison, other methods are also fine-tuned under the same experimental setting. Table~\ref{tab:sony} shows the quantitative results of our approach with existing SOTA on the Sony~\cite{TNTT} test sets. Our approach achieves the best results and the comprehensive PSNR outperforms the second place by 1.72dB. In Fig.~\ref{fig:real}, we test the fine-tuned model on real-world blurry videos and visually compare their deblurring and interpolation performance. As shown in the figure, our model restores accurate and sharp images from blurry videos, showing good generalization on real-world blurry sceneries.

\begin{table}
	\centering
	\caption{Quantitative comparisons on the Sony~\cite{TNTT} test sets for deblurring and multi-frame interpolation ($\times2$)).}
	\resizebox{9cm}{1.2cm}{
		\begin{tabular}{l|cc|cc|cc}
			\toprule
			\multirow{2}{*}{Method} &  \multicolumn{2}{c}{Deblurring}                                                                                                             & \multicolumn{2}{c}{Interpolation(x2)}                                                                                                      & \multicolumn{2}{c}{Comprehensiveness}                                                                                             \\
			\cline{2-7}
            & \multicolumn{1}{c}{PSNR} & \multicolumn{1}{c}{SSIM} & \multicolumn{1}{c}{PSNR} & \multicolumn{1}{c}{SSIM}           & \multicolumn{1}{c}{PSNR} & \multicolumn{1}{c}{SSIM} \\
			\hline
			BIN~\cite{BIN} & 38.81&0.9703&39.58&0.9740&39.20&0.9722   \\
			PRF~\cite{PRF} &39.15 &0.9720&40.28&0.9763&39.72&0.9742     \\
			DeMFI~\cite{DeMFI}  &38.76&0.9698&37.47&0.9681&38.12&0.9670   \\
			Ours & \textbf{42.22}&\textbf{0.9832}&\textbf{40.65}&\textbf{0.9775}&\textbf{41.44}&\textbf{0.9804} \\ 
			\bottomrule          
		\end{tabular}
	}
\label{tab:sony}
\end{table}

\subsection{Ablation Studies}
This section will discuss the influence of different submodules in our model. For fast evaluation, all the models are trained 300K iterations (about 50 epochs) on the $\times2$ BVFI task. Other training configurations are the same as our main experiments.

\subsubsection{Framework ablation}
This paper proposes an end-to-end three-stage framework to solve the BVFI problem. Here we will explore the effect of the order of the deblurring procedure and interpolation procedure on model performance. As shown in Table~\ref{tab:ablation}, the strategy of interpolation first and then deblurring achieves better performance than the other one. We consider that the deblurring procedure may eliminate some motion information hidden in blurry frames, which is bad for the subsequent interpolation process.

\begin{table}[t]
	\centering
	\caption{Ablation studies of our model with different submodules on the Adobe240~\cite{su2017deep} test set.}
	\resizebox{9cm}{2.8cm}{
		\begin{tabular}{l|c|cccc}
			\toprule
			Submodule Ablation&&\multicolumn{2}{c}{Deblurring}&\multicolumn{2}{c}{Interpolation}\\
			\hline
			\textbf{Framework}& Params(M)&PSNR & SSIM & PSNR &SSIM \\
			\hline
			Deblurring before Inter. &5.04&33.94&0.9398&34.53&0.9485  \\
			Inter. before deblurring(ours) &5.04&\textbf{34.88}&\textbf{0.9494}&\textbf{35.19}&\textbf{0.9557} \\
			\hline
			\textbf{Interpolation (ME)}& Params(M)&PSNR & SSIM & PSNR &SSIM \\
			\hline
			w/o ME &4.49&34.01 &0.9410&34.47&0.9494\\
			Optical flow&4.57 &34.31&0.9437&34.44&0.9488\\
			DConv (ours)&5.04&\textbf{34.88}&\textbf{0.9494}&\textbf{35.19}&\textbf{0.9557}\\
			\hline
			\textbf{The Nums of Taylor Order}& Params(M)& PSNR & SSIM & PSNR &SSIM \\
			\hline
			$n=1$ &5.04 &34.25&0.9433&34.65&0.9508\\
			$n=2$ (ours)&5.04 &34.88&0.9494&35.19&0.9557\\
			$n=3$&5.04 &\textbf{34.95} & \textbf{0.9499} &\textbf{35.32} &\textbf{ 0.9563}\\
			\hline
			\textbf{Deblurring Network}& Params(M)& PSNR & SSIM & PSNR &SSIM \\
			\hline
			ResNet &4.64&34.38&0.9439&34.68&0.9514\\
			UNet &5.20 &34.40&0.9443&34.68&0.9510\\
			Transformer (ours)&5.04&\textbf{34.88}&\textbf{0.9494}&\textbf{35.19}&\textbf{0.9557}\\
			\bottomrule          
		\end{tabular}
	}
	\label{tab:ablation}
\end{table}

\subsubsection{Temporal PCD module for frame interpolation}
In this paper, a temporal controllable PCD module is proposed to directly interpolate arbitrary multiple frames from blurry input frames. Following~\cite{EDVR}, we manually set the kernel size of the DConv layer as 3 and set the number of group as 8. It represents that $8\times3\times3=72$ offsets will be estimated for each pixel. 
To vilidate the usefulness of the TPCD module on the BVFI task, we compare our model with the model without motion estimation (ME) (as done in~\cite{BIN,ALANET,PRF}) and the model with optical flow motion estimation (as done in~\cite{TNTT,NIPS20,DeMFI}). The first strategy is the model without motion estimation. For a fair comparison, we retain the main structure of our TPCD module and regard the offset estimator network as an adaptive interpolation module to directly predict the intermediate features. The procedures for predicting the intermediate feature at $l$-th level can be formulated as:
\begin{equation}
	F_{0\rightarrow t}^l=\mathcal{F}_{E}([F_0^l,F_1^l,t], (F_{0\to t}^{l+1})^{\uparrow2}).
\end{equation}
The second strategy is optical flow-based motion estimation. For a fair comparison, we make a simple modification to the DConv layer in our TPCD module. Specifically, we set the group number and the kernel size as 1 and $1\times1$ respectively. In this way, the estimated deformable offsets can be regarded as the optical flows and the learned modulation masks can be considered as the occlusion masks.

Table~\ref{tab:ablation} shows the quantitative results. Since optical flows between blurry frames are highly uncertain, the optical flow-based model achieves a comparable performance with the model without ME. When we replace the optical flow with a DConv layer, the performance improves by 0.57dB and 0.75dB for the deblurring and interpolation tasks respectively. 

\begin{figure}[t]
	\centering
	\includegraphics[width=1\columnwidth]{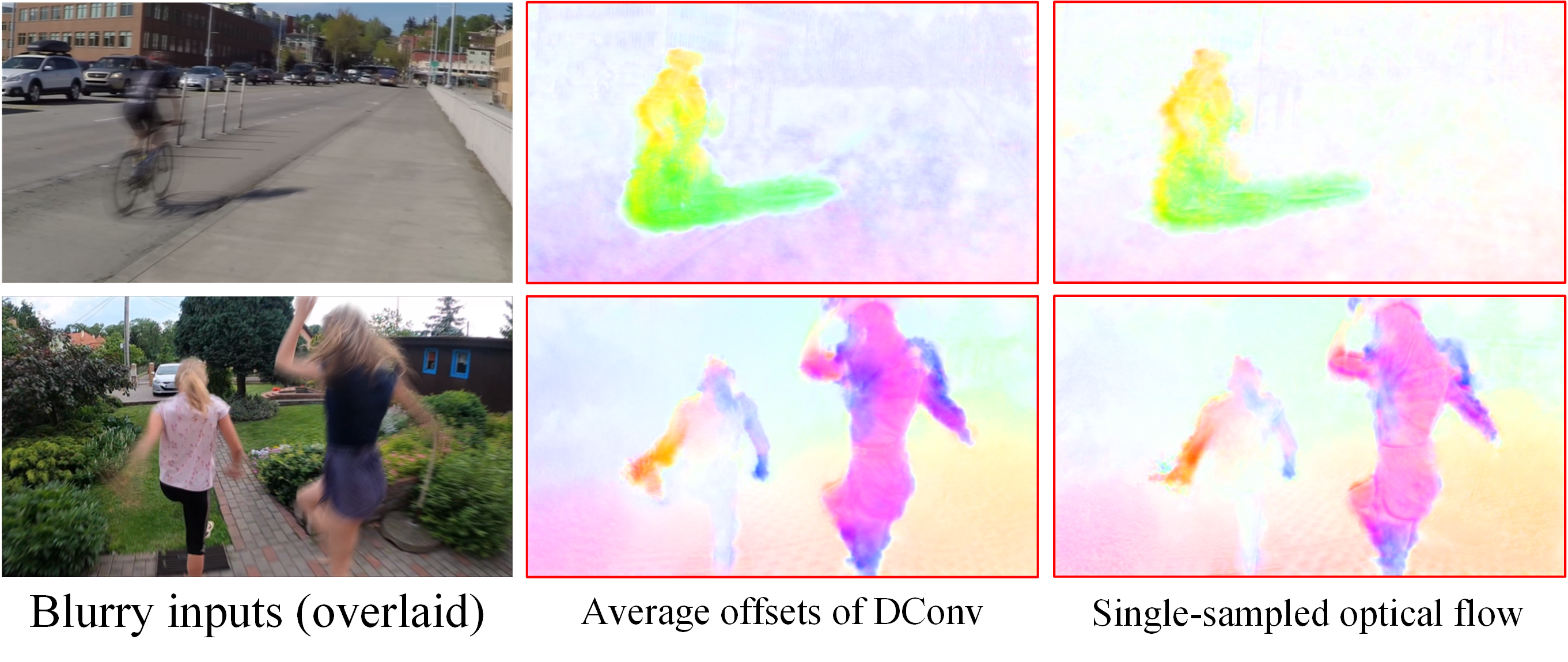} 
	\caption{The visualization of the learned deformable offsets (average) and the optical flow from blurry inputs.}
	\label{fig:deconv}
\end{figure}

\begin{figure}[t]
	\centering
	\includegraphics[width=0.9\columnwidth]{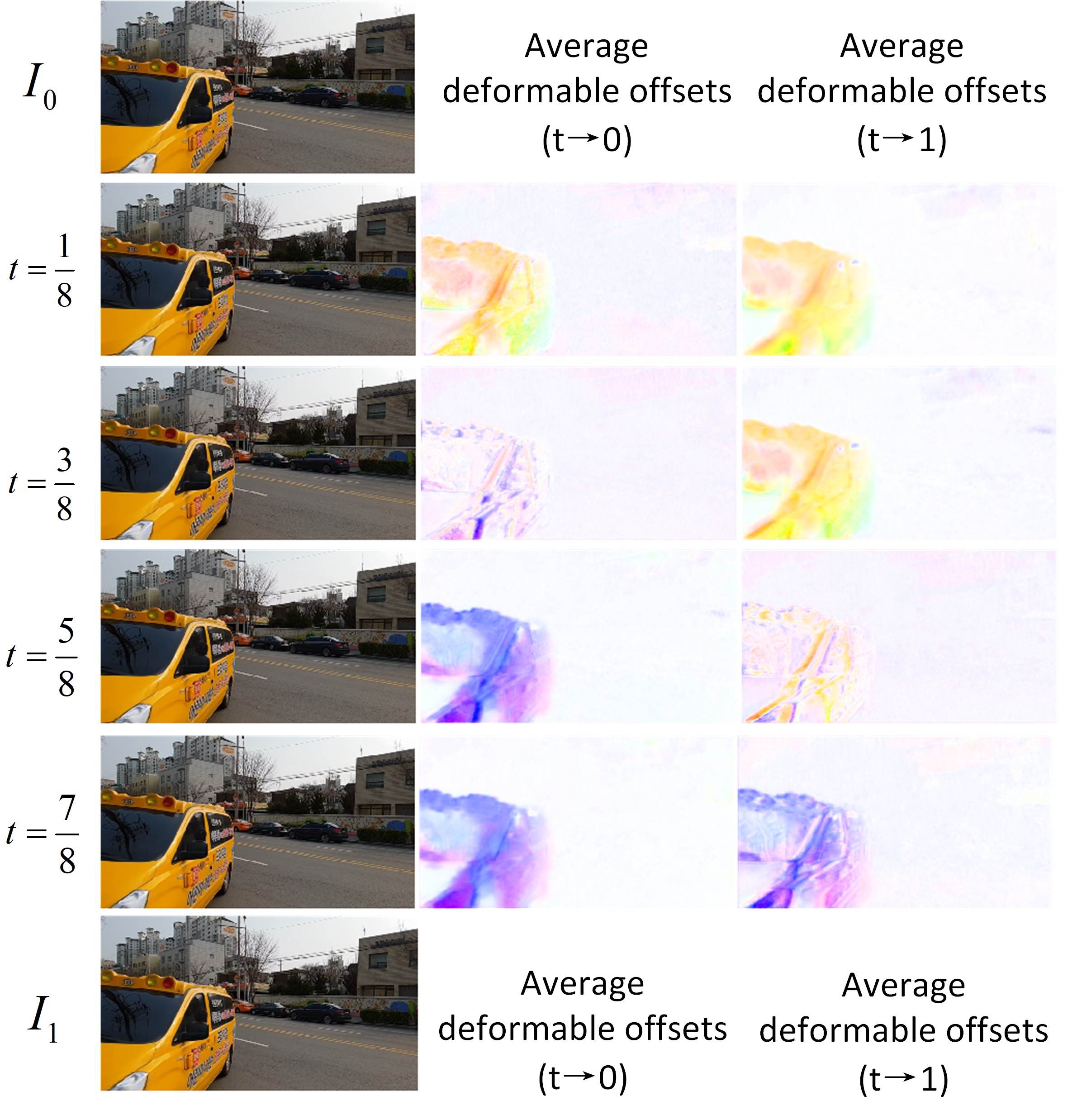} 
	\caption{The visualization of the temporally modulated average deformable offsets learned by the TPCD module.}
	\label{fig:flow_x8}
\end{figure}

In figure~\ref{fig:deconv}, we visualize the average offsets learned by the DConv layer and the single-sampled optical flow in $\times2$ BVFI task. Both two methods can estimate approximately accurate intermediate motions. However, the diverse sampling strategy helps the DConv layer learn more useful information from the blurry inputs, which can be proven by the significant performance improvement. In figure~\ref{fig:flow_x8}, we also visualize the temporally modulated average offsets learned by TPCD in $\times8$ BVFI task. As one can see, our method can effectively predict continuous intermediate flows.

\begin{figure}[t]
	\centering
	\includegraphics[width=1\columnwidth]{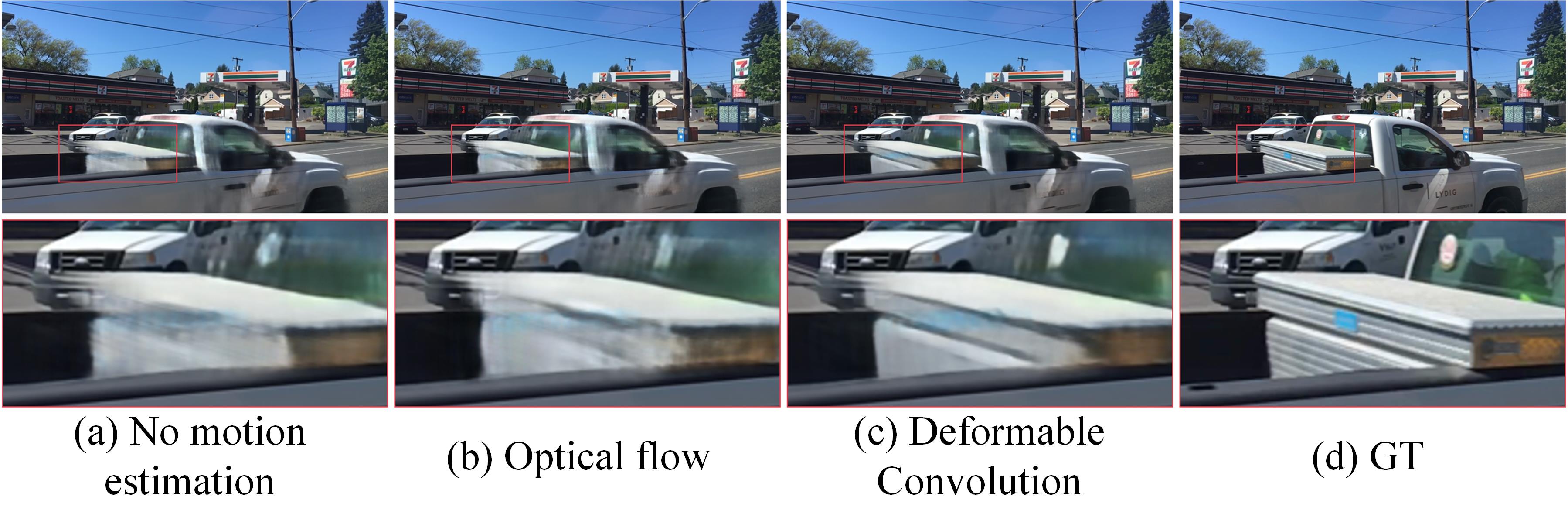} 
	\caption{Visual comparisons of our method with different motion estimation strategies.}
	\label{fig:ab_motion}
\end{figure}

In figure~\ref{fig:ab_motion}, we further compare their reconstruction results visually. From the figure, we can find that both the Non-motion estimation and optical flow-based motion estimation generate terrible results, while our deformable convolution can synthesize sharp edges from fast-moving cars. The experiments fully demonstrate the effectiveness of the adaptive DConv sampling for the BVFI task.


\begin{figure*}[t]
	\centering
	\includegraphics[width=1.7\columnwidth]{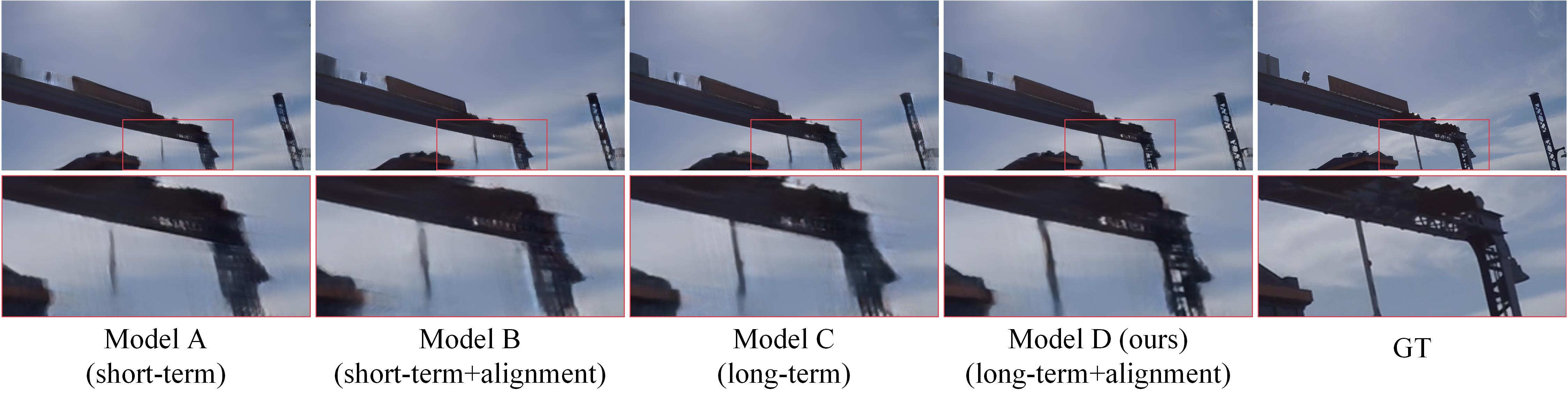} 
	\caption{Visual comparisons of our method with different temporal feature fusion strategies.}
	\label{fig:ab_fusion}
\end{figure*}

\subsubsection{The effect of the Bi-directional RDAM}
In this paper, we propose a Bi-directional RDAM to explore the long-term temporal information among multiple consecutive frames. Next, we will evaluate its effectiveness from two aspects: short-term fusion or long-term fusion, with or without feature alignment. The quantitative results are shown in Table~\ref{tab:temporal}. For the short-term model (A and B), we fuse the information among every three consecutive frames. For the long-term model (C and D), we fuse the information among all available frames. For the model without alignment, we remove the deformable alignment operations (as shown in Figure 5 in our main manuscript) and only use the adaptive fusion layer to fuse the temporal information. From Table~\ref{tab:temporal}, we can find that model B outperforms model A by 0.02dB and 0.07dB PSNR on deblurring and interpolation tasks. It indicates that the performance improvement of the alignment operation is limited when only short-term temporal information is considered. Model C outperforms model A by 0.15dB and 0.33dB on deblurring and interpolation tasks. When we introduce alignment operation in model C, the performance of model D improves by 0.83dB and 1.17dB PSNR on deblurring and interpolation tasks. The obvious performance improvement fully demonstrates the usefulness of the proposed long-term alignment feature fusion module. 

In figure~\ref{fig:ab_fusion}, we visually compare the reconstruction results of different models. Since our model effectively explores the long-term temporal information, it gets the best reconstruction results.

\begin{table}[t]
	\centering
	\caption{Ablation studies of our model with different temporal fusion strategies on the Adobe240 test set.}
	\resizebox{9cm}{1cm}{
		\begin{tabular}{c|ccc|cccc}
			\toprule
			\multirow{2}{*}{Model}&\multirow{2}{*}{Short-term}&\multirow{2}{*}{Long-term}&\multirow{2}{*}{Alignment}&\multicolumn{2}{c}{Deblurring}&\multicolumn{2}{c}{Interpolation}\\
			\cline{5-8}
			&&&&PSNR & SSIM & PSNR &SSIM \\
			\hline
			A&\checkmark&&&33.90&0.9396&33.69&0.9451 \\
			B&\checkmark&&\checkmark&33.92&0.9403&33.76&0.9460 \\
			C&&\checkmark&&34.05&0.9421&34.02&0.9466 \\
			D(ours)&&\checkmark&\checkmark&\textbf{34.88}&\textbf{0.9494}&\textbf{35.19}&\textbf{0.9557} \\
			\bottomrule          
		\end{tabular}
	}
	\label{tab:temporal}
\end{table}

\subsubsection{The effect of the number of the Taylor order}
\begin{figure}[t]
	\centering
	\includegraphics[width=1\columnwidth]{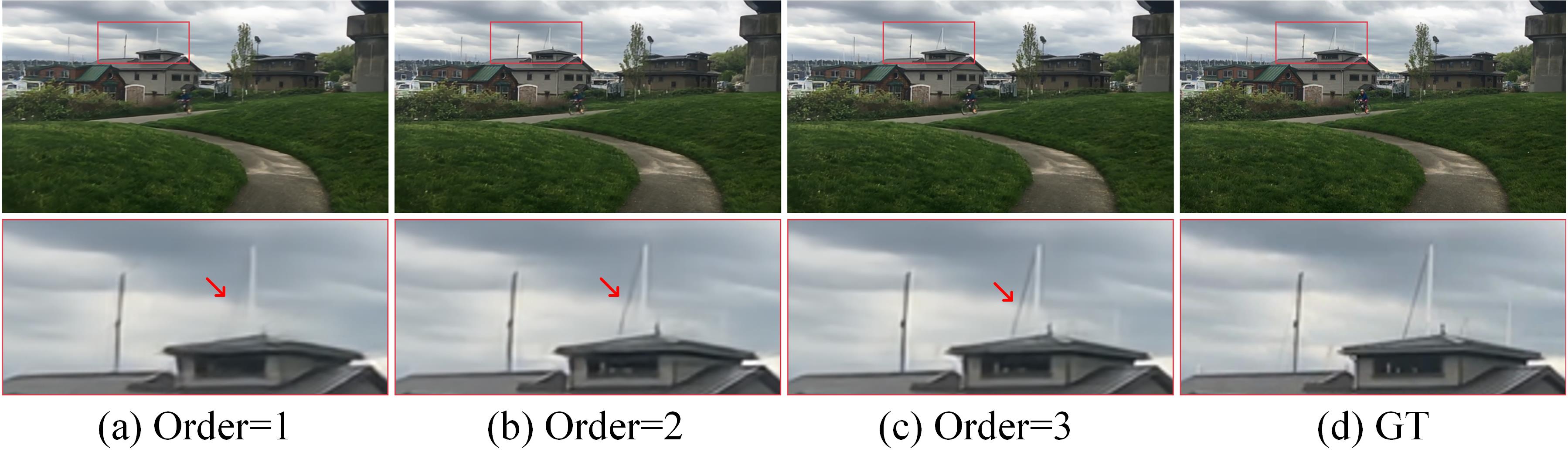} 
	\caption{Visual comparisons of our method with different Taylor orders.}
	\label{fig:ab_order}
\end{figure}

To show how the number of Taylor order $n$ affects the deblurring performance, we have compared the proposed method with different Taylor orders. Table~\ref{tab:ablation} shows the quantitative results of our model with $n\in[1,3]$. It can be observed that the higher order leads to better performance. From $n=1$ to $n=2$, the deblurring performance improves by 0.63dB. When we set $n=3$, the deblurring performance only improves by 0.07dB. Note that different recursions share the same network parameters in our model. Although the higher order will not increase the number of the parameter, it will introduce more computations. To balance the performance and the computational complexity, we manually set $n=2$ in our final model.
In figure~\ref{fig:ab_order}, we further visually compare their reconstruction results. It can be clearly seen that higher-order restores more details structures.

\subsubsection{The effect of the deblurring network} 
We simply compare the proposed transformer deblurring network with the ResNet and the modified UNet. To be specific, we employ 20 residual blocks~\cite{lim2017enhanced} to construct the ResNet. The modified UNet has the same structure as our transformer network, where the transformer layers are replaced by several “Conv” and “ReLU” layers. For a fair comparison, we control that the three models have a similar number of parameters. The quantitative results are shown in Table~\ref{tab:ablation}. Among these models, the proposed transformer deblurring network achieves the best performance.

\begin{figure}[t]
	\centering
	\includegraphics[width=1\columnwidth]{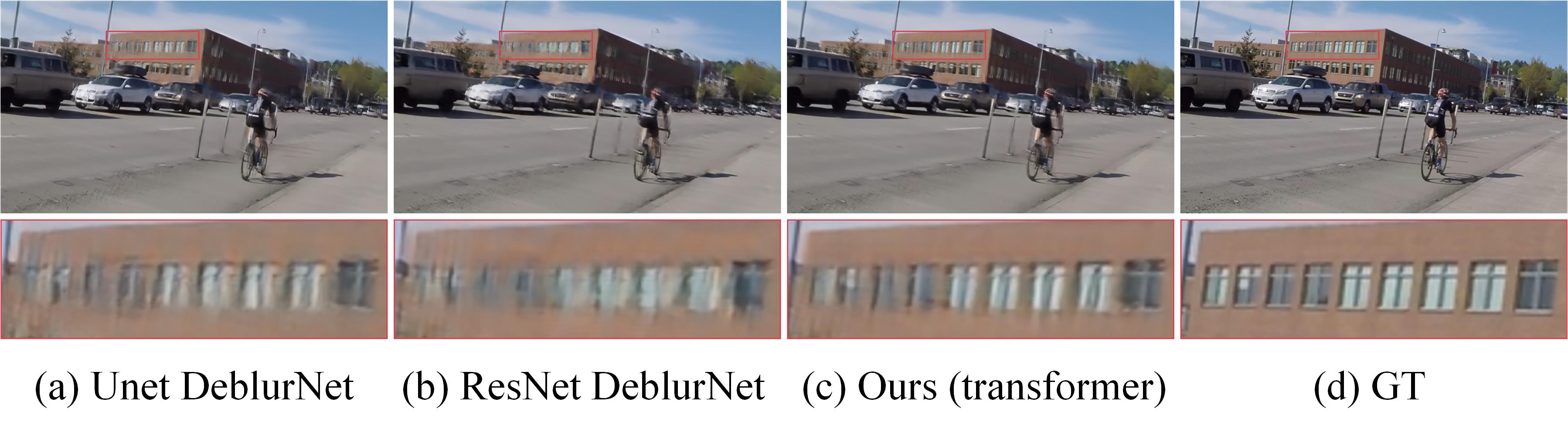} 
	\caption{Visual comparisons of our method with different deblurring networks.}
	\label{fig:ab_deblur}
\end{figure}

Figure~\ref{fig:ab_deblur} compares the results of our methods with different deblurring networks. Transformer has advantages in exploring long-range dependencies of an image. As shown in figure~\ref{fig:ab_deblur}, compared with ResNet and UNet deblurring networks, our transformer-empowered deblurring network can explore more global information, thus restoring more structural details.


\section{Conclusion}
In this paper, we have proposed an end-to-end three-stage BVFI framework to fully leverage all valuable information from blurry videos. Based on the deliberate thinking of the BVFI task, we decomposed the challenging problem into three subtasks, i.e., frame interpolation, temporal feature fusion and deblurring, and design specific network modules to handle these tasks, respectively. Compared with other BVFI methods, our three-stage framework can fully explore the hidden information in both intra-frame and inter-frame from input blurry videos. Since each module of our framework has clear task assignment, the framework also possesses good expandability. Experimental results demonstrate the effectiveness of the proposed method. In future work, we will focus on improving the efficiency of each submodule and designing real-time BVFI models.

\bibliographystyle{IEEEtran}     
\bibliography{Reference}

\end{document}